\documentclass[]{bytedance}
\usepackage[toc,page,header]{appendix}

\usepackage{minitoc}
\usepackage{amsfonts}
\usepackage{amssymb}
\usepackage{tabularx}
\usepackage{listings}
\usepackage{xcolor}
\usepackage{cancel}

\usepackage{tabulary,multirow,xspace}
\usepackage{fixmath,mathtools,nicefrac,mmstyle}
\usepackage{subcaption}
\usepackage{caption}
\usepackage{float}
\usepackage{wrapfig} %
\usepackage[misc]{ifsym} %
\usepackage{colortbl}

\usepackage{multicol}
\usepackage[most]{tcolorbox}
\usepackage{pifont}
\usepackage{amsmath}
\usepackage{amssymb}
\usepackage{graphicx}
\usepackage{makecell}
\usepackage{todonotes}
\usepackage{enumitem}
\usepackage{threeparttable}
\usepackage[table]{xcolor}
\usepackage{booktabs}
\usepackage{tabularx}
\usepackage{multirow}
\usepackage{subcaption}
\usepackage{wrapfig}
\usepackage{comment}
\usepackage{array}
\usepackage{rotating}
\usepackage{adjustbox}
\usepackage{listings}

\definecolor{codegreen}{rgb}{0,0.6,0}
\definecolor{codegray}{rgb}{0.5,0.5,0.5}
\definecolor{codepurple}{rgb}{0.58,0,0.82}
\definecolor{backcolour}{rgb}{0.95,0.95,0.92}
\definecolor{boxblue}{RGB}{57,89,163}
\definecolor{boxbluebg}{RGB}{230,237,250} 

\lstdefinestyle{mystyle}{
    backgroundcolor=\color{backcolour},   
    commentstyle=\color{codegreen},
    keywordstyle=\color{magenta},
    numberstyle=\tiny\color{codegray},
    stringstyle=\color{codepurple},
    basicstyle=\ttfamily\footnotesize,
    breakatwhitespace=false,         
    breaklines=true,                 
    captionpos=b,                    
    keepspaces=true,                 
    numbers=none,                    
    numbersep=5pt,                  
    showspaces=false,                
    showstringspaces=false,
    showtabs=false,                  
    tabsize=2
}
\definecolor{mygray1}{gray}{.95}
\definecolor{mygray2}{gray}{.9}
\definecolor{mygray3}{gray}{.95}
\usepackage{pifont}%

\newlength\savewidth
\newcolumntype{x}[1]{>{\centering\arraybackslash}p{#1pt}}

\newcommand{\app}{\raise.17ex\hbox{$\scriptstyle\sim$}}

\usepackage{xcolor}
\usepackage{listings}

\definecolor{commentgreen}{rgb}{0.1, 0.4, 0.1}
\definecolor{keywordblue}{rgb}{0.1, 0.1, 0.7}
\definecolor{stringred}{rgb}{0.7, 0.1, 0.1}

\lstdefinestyle{mystyle}{
    commentstyle=\color{commentgreen},
    keywordstyle=\color{keywordblue},   
    stringstyle=\color{stringred},
    basicstyle=\ttfamily\scriptsize, 
    breaklines=true,
    keepspaces=true,
    showstringspaces=false,
    frame=none,                     
    language=Python, 
}

\newcommand{\name}{Bernini}
\title{\name{}: Latent Semantic Planning for Video Diffusion}

\author{
\centerline{
Bernini Team, Bytedance
}
}

\abstract{
Multimodal large language models (MLLMs) and diffusion models have each reached remarkable maturity: MLLMs excel at reasoning over heterogeneous multimodal inputs with strong semantic grounding, while diffusion models synthesize images and videos with photorealistic fidelity. We argue that these two families can be unified through a simple division of labor: MLLMs perform semantic planning, while diffusion models render pixels from high-level semantic guidance and low-level visual features. Building on this idea, we propose \textbf{Bernini}, a unified framework for video generation and editing. An MLLM-based \textbf{planner} predicts the target semantic representation directly in the ViT embedding space, and a DiT-based \textbf{renderer} synthesizes pixels conditioned on this plan, augmented by text features and, for editing, source VAE features for detail preservation. Because semantics serve as the interface, the planner and renderer can be trained separately and only lightly co-trained, preserving the pretrained strengths of both components while keeping training efficient. To better handle multiple visual inputs, we introduce Segment-Aware 3D Rotary Positional Embedding (SA-3D RoPE), and further incorporate chain-of-thought reasoning in the planner to better transfer understanding into generation. Bernini achieves state-of-the-art performance across a wide range of video generation and editing benchmarks, and is, to our knowledge, the first open-source model to approach leading closed-source systems on video editing.%
}

\date{\today} \correspondence{\email{zhyuan001@gmail.com}}
\checkdata[Project Page]{\url{https://bernini-ai.github.io}}

\begin{document}
\maketitle
\vspace{-20pt}
\begin{figure}[!ht]
    \centering
    \includegraphics[width=\linewidth]{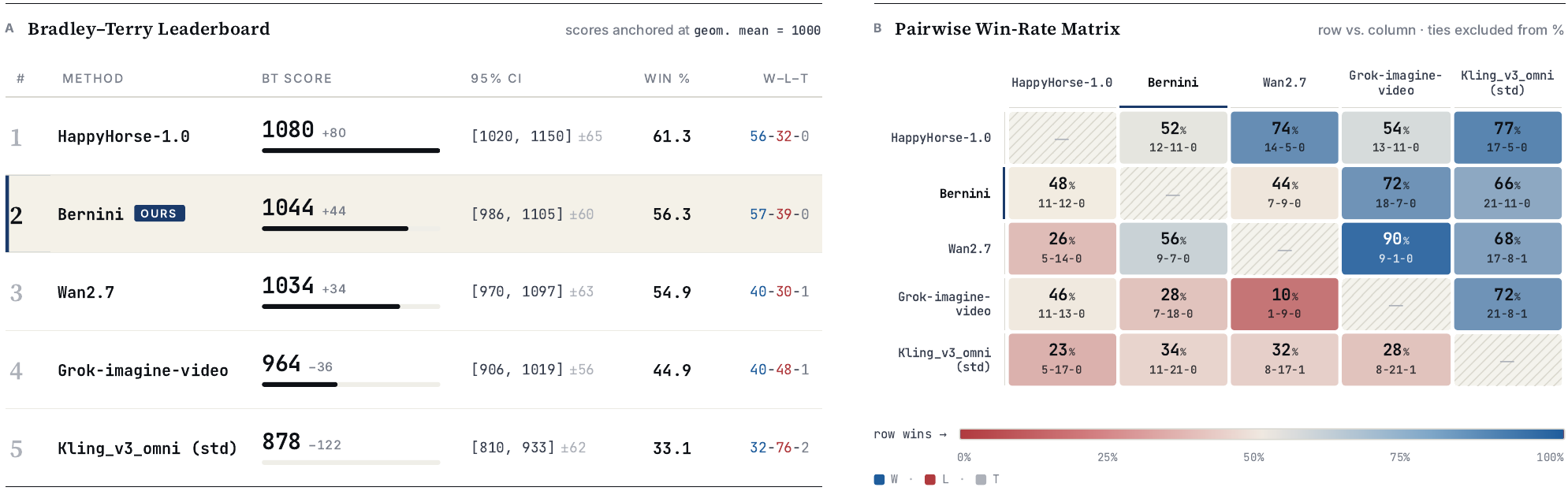}
\vspace{-18pt}
\caption{
  \textbf{Video editing leaderboard.} Pairwise human preferences on open-ended video editing (no restriction on instruction type).
  \textbf{(A) Leaderboard.} BT scores~\cite{bradley1952rank} with $95\%$ bootstrap CIs; \textbf{Win \%}\,$=1/(1+10^{-(s-1000)/400})$ vs.\ avg.\ opponent;
  \textbf{W-L-T} raw record.
  \textbf{(B) Win-rate matrix.} $W/(W{+}L)$;
  Bernini: 480p/24\,fps; baselines: 720p/24\,fps.
  Details in Appendix~\ref{app:editing-arena}.
}
\label{fig:teaser0}
\end{figure}

\begin{figure}[h]
    \centering
    \includegraphics[width=\linewidth]{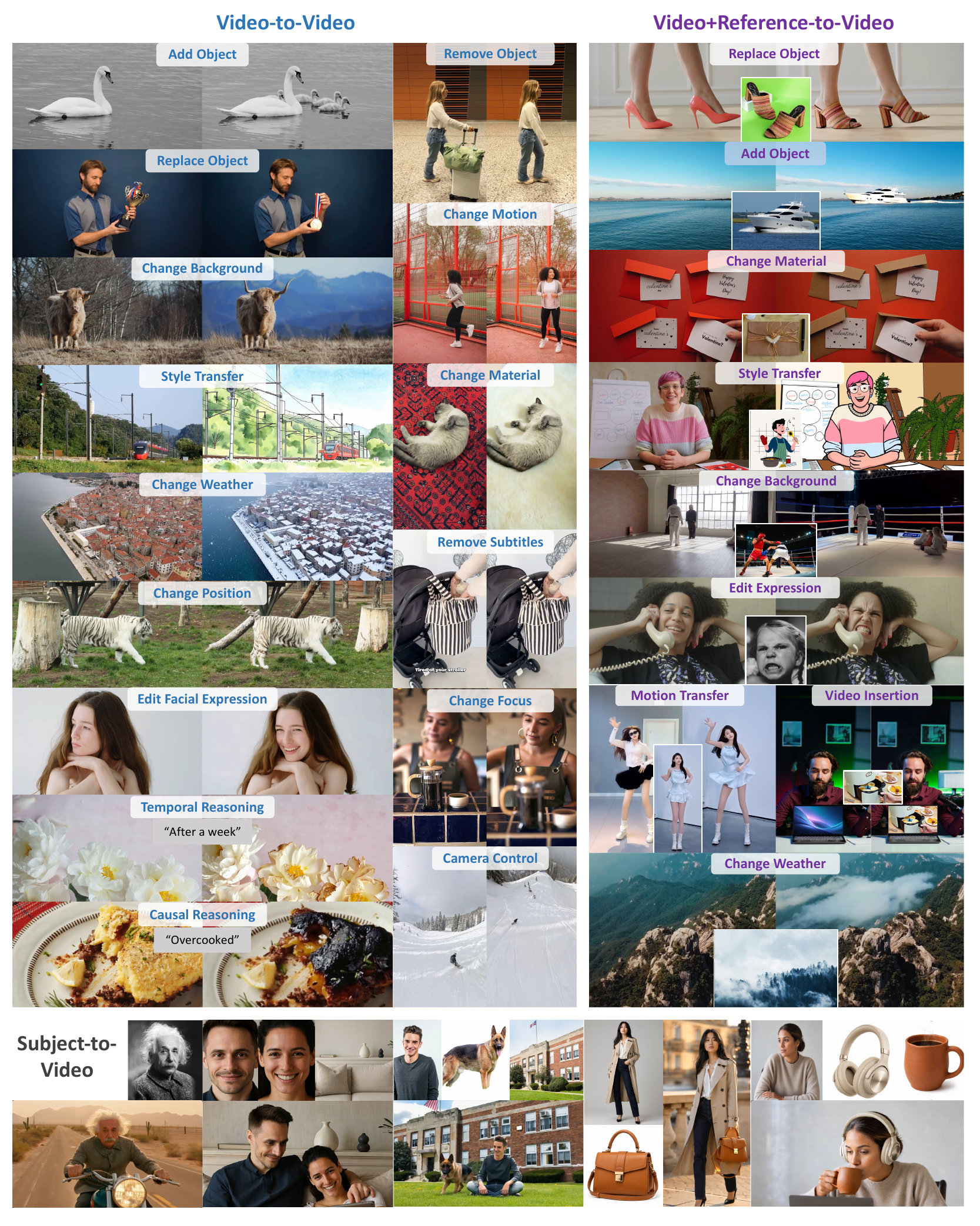}
    
\caption{Bernini supports diverse video generation tasks within a unified framework, including text-to-video (T2V), reference-to-video (R2V), video editing (V2V), and reference-guided video editing (RV2V).}
\label{fig:teaser2}
\end{figure}

\clearpage

\tableofcontents
\clearpage

\section{Introduction}
Multimodal large language models (MLLMs)~\cite{qwen2.5vl,internvl,llava} and diffusion models~\cite{stablediffusion,flux,sd3,sora,wan} have matured along largely independent trajectories. Modern MLLMs read long instructions, reason over multiple reference images, and ground their answers in a complex multimodal context. Diffusion models, meanwhile, have become the default tool for photorealistic image and video synthesis at high resolutions and long durations. %
The natural next step is to combine these two mature families into a single system that both understands intent and generates the desired output, supporting unified understanding, generation, and editing within one model. However, how to do so effectively remains an open question.

Our approach begins with two simple observations. First, MLLMs are naturally suited to \emph{semantic reasoning}: interpreting long instructions, grounding on multiple references, and forming an internal representation of what the output should be. Second, diffusion generation decomposes cleanly into \emph{semantic guidance} and \emph{detail preservation}. The high-level content is determined by a compact semantic signal, while fine-grained fidelity, and in editing also consistency with the source input, requires dense pixel-level latents such as VAE features. Crucially, the semantic signal itself need not be high-resolution to be effective. A handful of semantic tokens are enough to specify an entire scene.

These observations suggest a clean division of labor: let the MLLM carry out semantic reasoning, and let the diffusion model focus on synthesis, using semantic features as its primary condition and pixel-level features only where detail preservation demands them. A natural question is what representation should carry the semantic signal between the two. We anchor this interface to a representation that already exists within MLLM itself, namely its own \emph{ViT embedding space}~\cite{vit,radford2021learning,siglip}. The MLLM already reasons and represents visual content in this space, so training it to plan the target in ViT embeddings aligns naturally with its pretrained representations and requires minimal adaptation.

We instantiate this principle as \textbf{Bernini}, a unified framework for multimodal understanding, generation, and editing. Bernini consists of a \textbf{planner}, based on an MLLM, that predicts the target's visual representation in the continuous ViT embedding space. Following a masked generative modeling paradigm~\cite{li2024autoregressive}, a lightweight ViT embedding decoder on top of the MLLM recovers randomly masked target ViT tokens from the hidden states of the MLLM, and at inference progressively fills in the full target representation from fully masked tokens. The \textbf{renderer}, a Diffusion Transformer (DiT)~\cite{dit}, then synthesizes the final image or video through flow-matching~\cite{flowmatching} denoising over VAE latent tokens, conditioned on the semantic embedding from the planner through cross-attention and augmented with text features. For editing tasks, VAE features of the source input are additionally injected to preserve detail and consistency.

To unify different task types, we adopt a shared input protocol across text-to-video, reference-to-video, and editing, achieving broad modality coverage without task-specific architectures. For multiple visual sources within a unified sequence, we further introduce \textbf{Segment-Aware 3D Rotary Positional Embedding (SA-3D RoPE)}, which augments standard spatiotemporal rotary embeddings~\cite{rope} with a segment-index-conditioned phase modulation. 
Finally, to amplify the contribution of understanding to generation, the planner is equipped with a \textbf{Chain-of-Thought (CoT) mechanism}~\cite{cot,visualsketchpad} that performs reasoning in latent space before producing the final embedding. 
Because semantics serve as the interface, the two components can be trained largely independently and only lightly co-trained thereafter, preserving the MLLM's pretrained capabilities and allowing its multimodal understanding to transfer directly into diverse downstream generation tasks. Our contributions are summarized as follows:
\vspace{-3pt}
\begin{itemize} [leftmargin=*,itemsep=1pt,topsep=2pt]
    \item We propose \textbf{Bernini}, a unified framework for generation, and editing that uses the MLLM's own ViT embedding space as a semantic bridge to the diffusion generator, allowing pretrained understanding to transfer directly into generation and enabling strong generalization across diverse video tasks.
    \vspace{-3pt}
    \item We design a suite of data construction pipelines that yield a large-scale, multi-task corpus for unified video generation and editing, spanning video- and image-pair pretraining data, high-quality propagation-based and motion-aware video editing data, and reference-image- and reference-video-guided generation data, providing the diverse and high-fidelity supervision required to train Bernini across all tasks.
    \vspace{-3pt}
    \item Bernini achieves state-of-the-art performance across a wide range of video generation, editing, and reference-to-video benchmarks, including OpenVE-Bench, OpenS2V-Eval, and our newly proposed Bernini-Bench.
\end{itemize}

\begin{figure}[t]
    \centering
    \includegraphics[width=\linewidth]{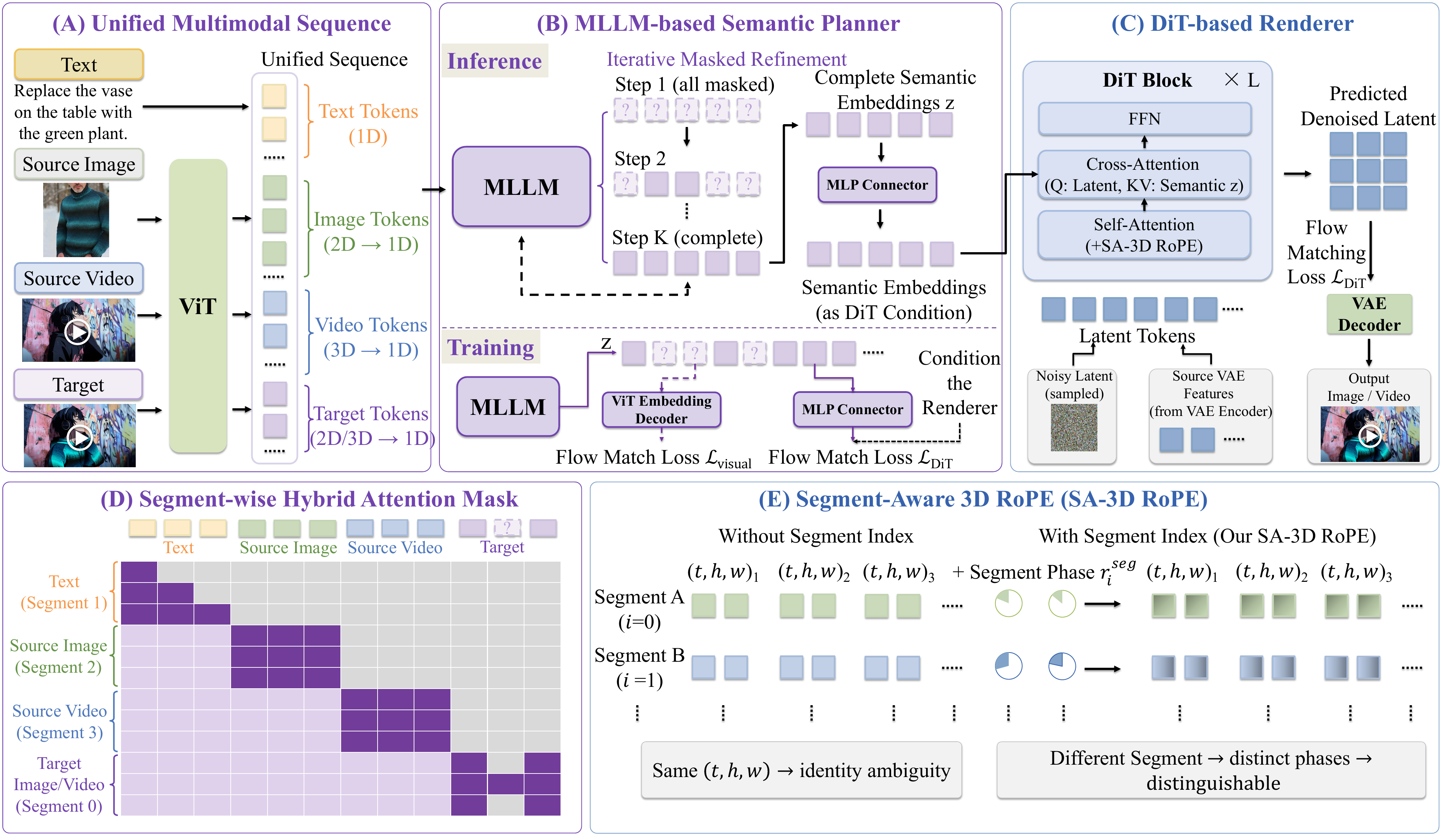}
    \caption{Overview of Bernini. (A) Visual and text inputs are serialized into a unified 1D sequence. (B) The MLLM planner predicts target semantic embeddings from masked targets and conditions the renderer. (C) The DiT-based renderer performs flow matching in the VAE latent space conditioned on semantic embeddings and source VAE features. (D) Bernini uses a segment-wise hybrid attention mask in MLLM. (E) Segment-aware 3D RoPE disambiguates visual tokens from different segments.}
    \label{fig:framework}
\end{figure}

\section{Methods}

\subsection{Architecture}
As illustrated in Fig.~\ref{fig:framework}, Bernini consists of two main components: an MLLM-based planner and a DiT-based renderer. 
Taking multimodal conditions as input, the MLLM performs multimodal understanding and semantic reasoning to produce the desired target content.
An MLP connector then maps these hidden states into the conditioning representation required by the DiT-based renderer. Conditioned on these semantic features, together with additional text features and source visual conditions when available, the DiT-based renderer synthesizes the final image or video in the VAE latent space.

\subsubsection{MLLM-based Planner}
\textbf{Unified Input Formulation.}
To support diverse tasks within a single framework, Bernini adopts a unified multimodal input formulation. All task instances, including text-to-video generation, text-to-image generation, reference-to-video generation, and image or video editing, are serialized into a shared token sequence composed of textual tokens and visual tokens from the source inputs and the target output.
Formally, given a multimodal input sequence, the MLLM encodes the entire sequence and produces contextualized hidden states \(\mathbf{z}\) that capture the target intent conditioned on the input context:
\begin{equation}
\mathbf{z} = \mathrm{MLLM}(\mathbf{t}, \mathbf{v}^{\text{src}}_1, \mathbf{v}^{\text{src}}_2, \dots, \mathbf{v}^{\text{src}}_N, \mathbf{v}^{\text{tgt}})
\label{eq:mllm}
\end{equation}
where \(\mathbf{t}\) denotes the input textual embeddings, \(\mathbf{v}^{\text{src}}_i\) denotes the ViT embeddings of the \(i\)-th source visual input, \(N\) is the number of source inputs, and \(\mathbf{v}^{\text{tgt}}\) denotes the visual embeddings corresponding to the target output. During training, \(\mathbf{v}^{\text{tgt}}\) is partially masked at random, while during inference it is initialized as fully masked.

\textbf{Mask-based Semantic Planning.}
Motivated by the intrinsically bidirectional nature of visual semantic latents, a masked generative modeling paradigm~\cite{he2026vidlada,chang2022maskgit} is adopted to better capture contextual dependencies. To mitigate the visual information loss introduced by discrete tokenization, we represent visual tokens as dense embeddings, which serve as both the input and output of the MLLM.

During training, a subset of target visual tokens is randomly masked and replaced with a shared mask token. The masking ratio is sampled from a Beta distribution,
\(r \sim \mathrm{Beta}(\alpha, \beta)\),
where \(\alpha\) and \(\beta\) are hyperparameters. The MLLM is then trained to infer the masked content from the remaining visible tokens together with the surrounding multimodal context. The resulting hidden states serve as semantic embeddings for the target visual content.
To recover the target ViT embeddings from these semantic embeddings, we follow the design philosophy of MAR~\cite{li2024autoregressive}. Specifically, the hidden states at masked positions are fed into the ViT embedding decoder, which consists of an MLP followed by a ResNet-based prediction head. The decoder predicts the corresponding ground-truth ViT embeddings and is trained with a flow-matching objective in the ViT embedding space.

During inference, all target visual tokens are initialized as masked tokens. The MLLM then progressively decodes the target representation over $K$ refinement steps, following the standard masked generative paradigm. At step $k$
, the mask ratio is scheduled as
\(
\mathrm{mask\_ratio}(k, K) = \cos\left(\frac{\pi}{2}\cdot\frac{k+1}{K}\right)
\),
so that the number of masked tokens gradually decreases over time. At each step, the currently predicted tokens are fed back into the MLLM and used as partial observations for the next round of prediction. This iterative process progressively refines the target representation from coarse to fine, until a complete target ViT embedding sequence is obtained.

\subsubsection{DiT-based Renderer}
The DiT-based renderer performs diffusion in the VAE latent space, using the contextualized hidden states \(\mathbf{z}\) from the MLLM in Eq.~\ref{eq:mllm} as conditioning features, and decodes the resulting target latent into the final output.
In addition, VAE features extracted from the source image or video are incorporated to preserve low-level details and ensure consistency with the source content.

\textbf{Segment-Aware 3D RoPE.}
In DiTs, 3D RoPE is commonly used to encode temporal and spatial positions for visual tokens.
It encodes the temporal, vertical, and horizontal positions of each visual token into three rotary subspaces and concatenates them to form \(\mathbf{r}_{t,h,w}\).
When Bernini concatenates all visual inputs and output as a unified sequence, 
tokens from different segments (different reference images, source videos, or target output) may share the same \((t,h,w)\) coordinates, making it difficult to distinguish different identities.
To address this issue, SA-3D RoPE is introduced, which assigns each visual segment an index \(i\), e.g. \(i=0\) for the target segment and \(i=1,2,\dots\) for input segments, and 
incorporates the segment index directly into the rotary position encoding.
To be specific, 
a full-dimensional rotary frequency vector \(\mathbf{r}^{\mathrm{seg}}_i\) is constructed to additionally encode the segment index \(i\) for each segment index.
Then, SA-3D RoPE can be calculated through
\begin{equation}
\tilde{\mathbf{r}}_{t,h,w,i} = \mathbf{r}_{t,h,w} \odot \mathbf{r}^{\mathrm{seg}}_i,
\end{equation}
where \(\odot\) denotes multiplication of complexes in element order.
This introduces a segment-dependent global phase modulation on top of the original spatiotemporal phase. On the one hand, it allows attention to distinguish tokens from different segments while preserving the original spatial-temporal modeling properties of 3D RoPE. On the other hand, because the segment phase is a continuous function of $i$, it is also readily scalable, since segment indices beyond the trained range
can be evenly interpolated into the trained range at inference, supporting more reference inputs than seen during training without retraining
(see Appendix~\ref{app:src_id_interp}).

\subsection{Training Objectives}
During training, the MLLM is optimized with the standard next-token prediction (NTP) loss $\mathcal{L}_{\mathrm{ntp}}$ to preserve its multimodal understanding capability. The ViT embedding decoder and the DiT renderer are both trained with standard flow-matching objectives, denoted as $\mathcal{L}_{visual}$ and $\mathcal{L}_{dit}$ in the continuous ViT embedding space and the VAE latent space, respectively. The two objectives share the same formulation, differing only in the definition of the target representation and the corresponding velocity field. The overall training objective is the weighted sum of these three losses:
\begin{equation}
\label{eq:total_loss}
\mathcal{L}
=
\lambda_{\mathrm{text}} \mathcal{L}_{\mathrm{ntp}}
+
\lambda_{\mathrm{visual}} \mathcal{L}_{\mathrm{visual}}
+
\lambda_{\mathrm{dit}} \mathcal{L}_{\mathrm{dit}},
\end{equation}
where \(\lambda_{\mathrm{text}}\), \(\lambda_{\mathrm{visual}}\), and \(\lambda_{\mathrm{dit}}\) are the corresponding loss weights.

\section{Data}

Bernini is trained on a diverse corpus that includes text-only, multimodal understanding, image/video generation, and image/video editing tasks. 
Although substantial progress has been made in constructing understanding data~\cite{wiedmann2025finevision,zhang2024llava}, image editing data~\cite{zhang2023magicbrush, chen2025sharegpt4oimage, wei2024omniedit, ye2025imgedit, kuprashevich2025nohumansrequired, zhao2024ultraedit, yu2025anyedit, wang2025gptimageedit}, and a limited amount of video editing data~\cite{bai2025recammaster, zi2025senorita, luo2025camclonemaster} has also been explored, the current landscape remains insufficient for training general-purpose video editing models. Video editing spans diverse task types, yet mature and scalable data construction pipelines are still lacking. In addition to incorporating existing open-source data into our training corpus, we further explore a series of data construction strategies for both large-scale pretraining and high-quality supervised fine-tuning, including video-to-video editing, reference-image-based video generation and editing, reference-video-based video generation, and reasoning-augmented video data.

\subsection{Pre-training Data}
\textbf{Video-pair Data.}
Current video editing models are constrained by the limited scale and quality of available training data, as existing video editing datasets are often noisy and rely on immature construction pipelines. This challenge makes large-scale video-pair pre-training essential. To this end, we constructed a large-scale dataset comprising 20 million video pairs from general T2V corpora. Our pipeline constructs diverse and balanced video pairs from raw videos through similarity-based filtering, content-aware sampling, and coarse-to-fine instruction generation. 

\begin{figure}[t]
\centering %

\begin{subfigure}[b]{0.55\linewidth} %
    \centering
    \includegraphics[width=\linewidth]{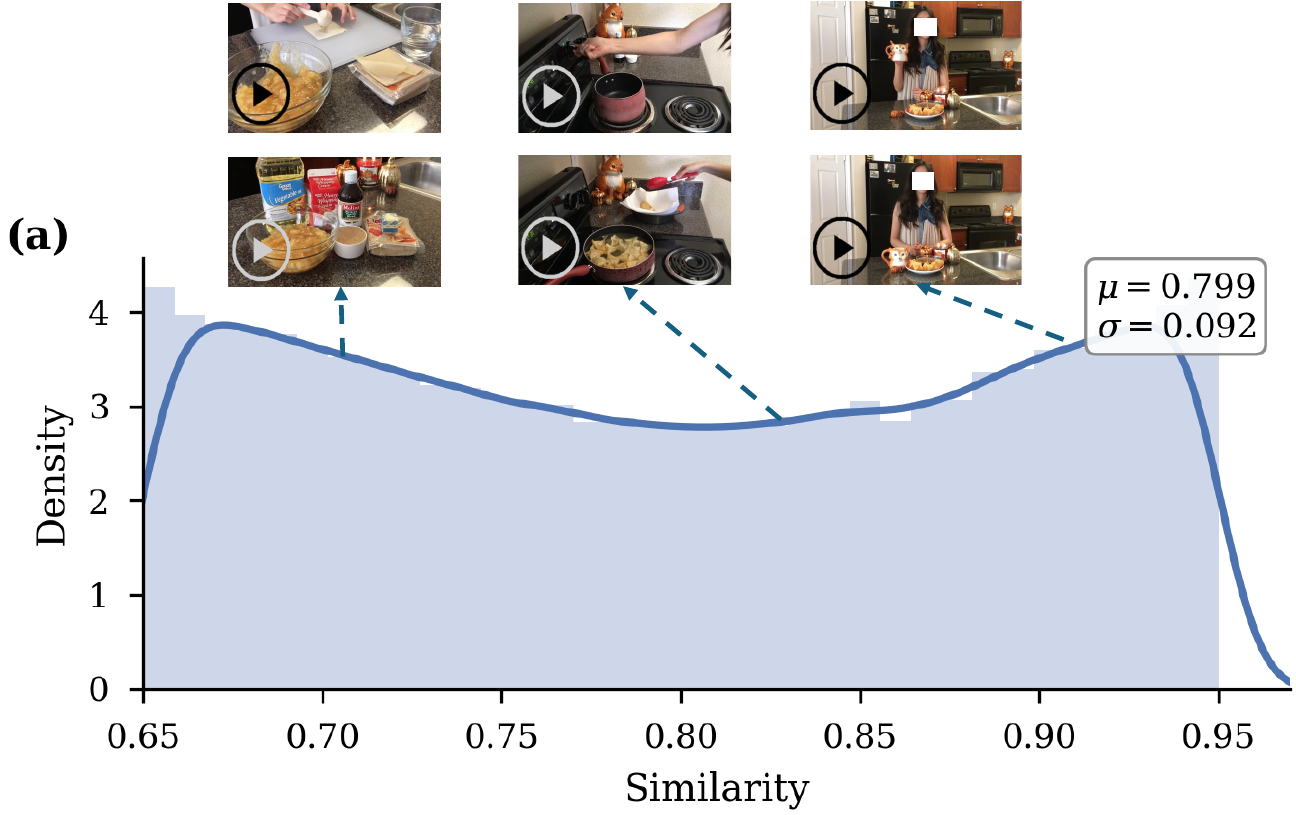}
\end{subfigure}%
\hfill %
\begin{subfigure}[b]{0.43\linewidth} %
    \centering
    \includegraphics[width=\linewidth]{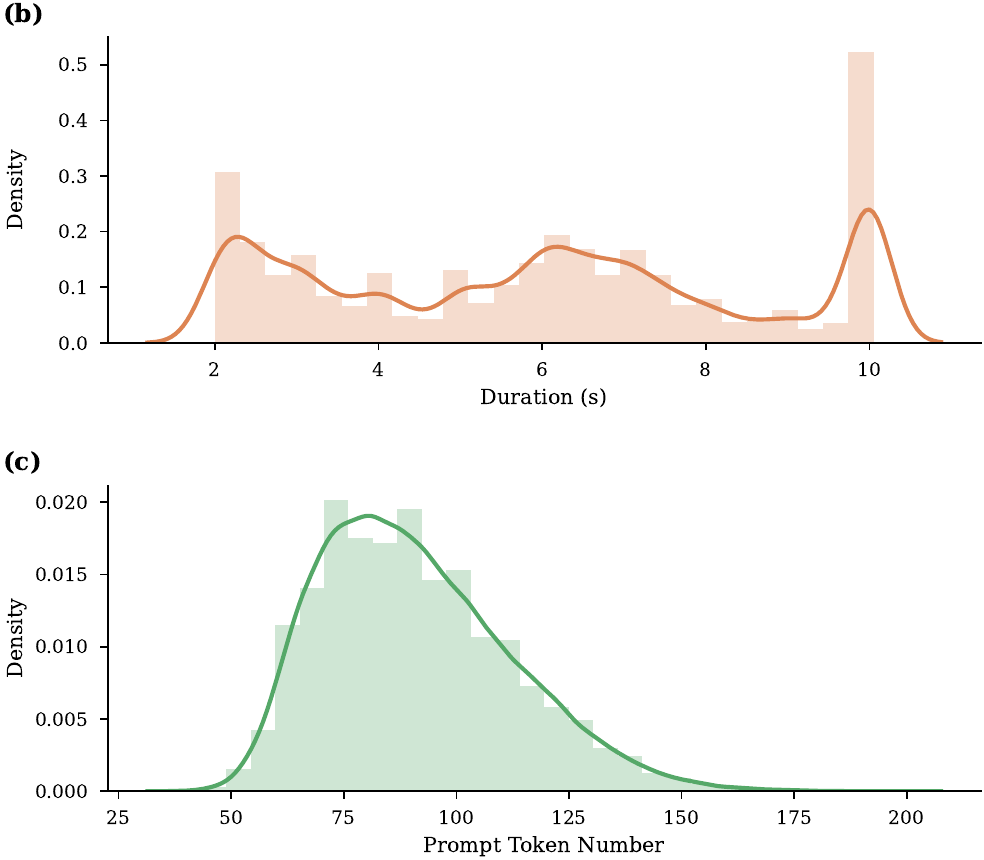}
\end{subfigure}

\caption{Dataset statistics of the collected video pairs. (a) Similarity score distribution between 0.65 and 0.95, with representative video pairs, extracted from the same raw video, visualized at their corresponding scores. (b) Distribution of video durations. (c) Distribution of prompt token counts.}

\label{fig:dataset_stats_video_pair}

\end{figure}

\begin{figure}[htbp]
    \centering
    \includegraphics[width=1.\linewidth]{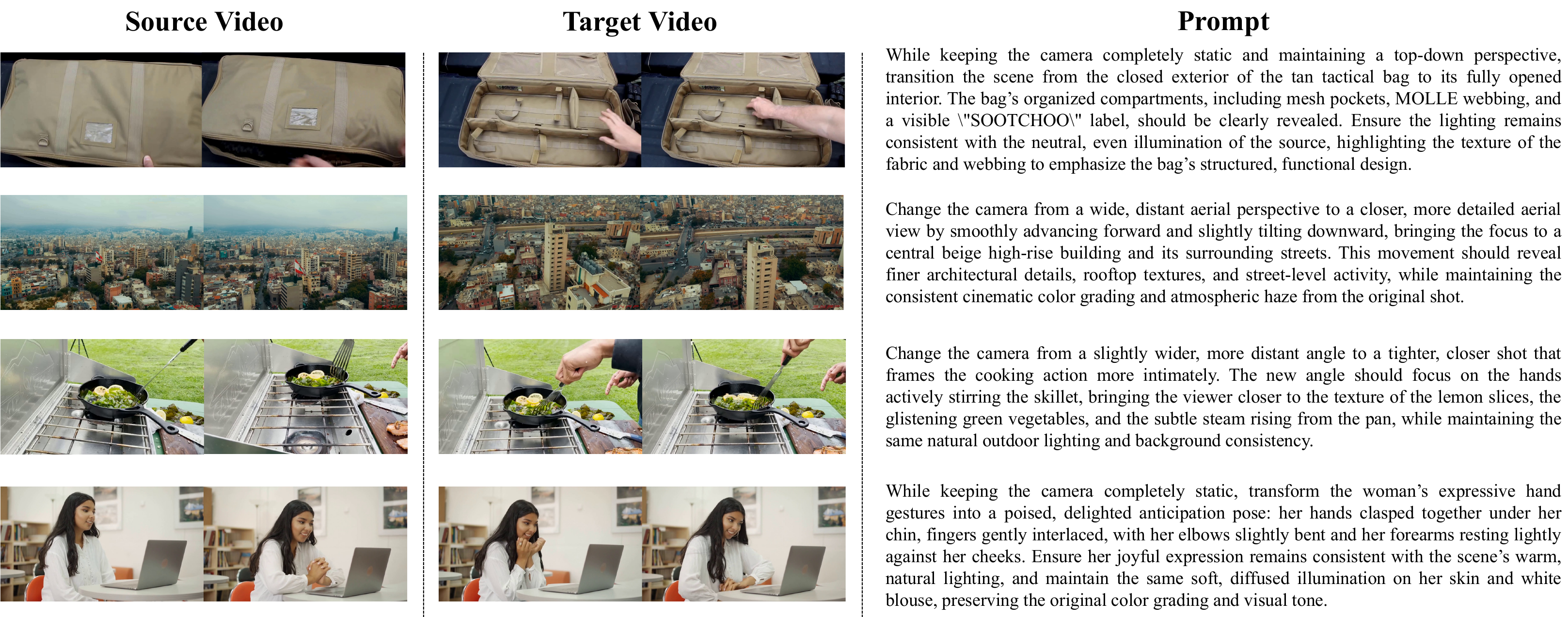}
    \caption{Examples of video pairs extracted from general T2V corpora, along with their generated dense prompts.}
    
    \label{fig:video_pair_case}
\end{figure}

Specifically, for video clips originating from the same raw video, we compute their global representations using X-CLIP~\cite{ma2022x} and compute pairwise similarity scores between video pairs. 
The selected video pairs that jointly satisfy the following conditions: (1) a similarity score between 0.65 and 0.95; (2) a duration between 2 and 10 seconds; 
(3) a 1:1 ratio of human-centric to non-human-centric content, which is annotated by Qwen3-VL-30B-A3B-Instruct~\cite{bai2025qwen3}; and 
(4) limiting each raw video to a maximum of 100 pairs. This approach ensures a balance between spatio-temporal coherence and content variety. 
Finally, to generate high-quality instructional prompts, we employ Qwen3-VL-235B-A22B-Instruct~\cite{bai2025qwen3} using a coarse-to-fine strategy. This approach first generates a coarse transition description between the video clips, which is subsequently refined into a detailed prompt. This enables fine-grained descriptions of camera motion as well as changes in the foreground and background.

Figure~\ref{fig:dataset_stats_video_pair} presents the statistical distributions of our collected video pairs, including similarity scores, video durations, and generated prompt token counts. The similarity scores are approximately uniformly distributed, while the video durations and prompt lengths span a wide spectrum, demonstrating the overall diversity of our dataset. Furthermore, Fig.~\ref{fig:video_pair_case} shows examples of the constructed video pairs alongside their corresponding generated dense prompts. Each prompt is structured to first detail the camera motion, followed by descriptions of the foreground and background changes.

\begin{figure}[t]
    \centering
    \includegraphics[width=\linewidth]{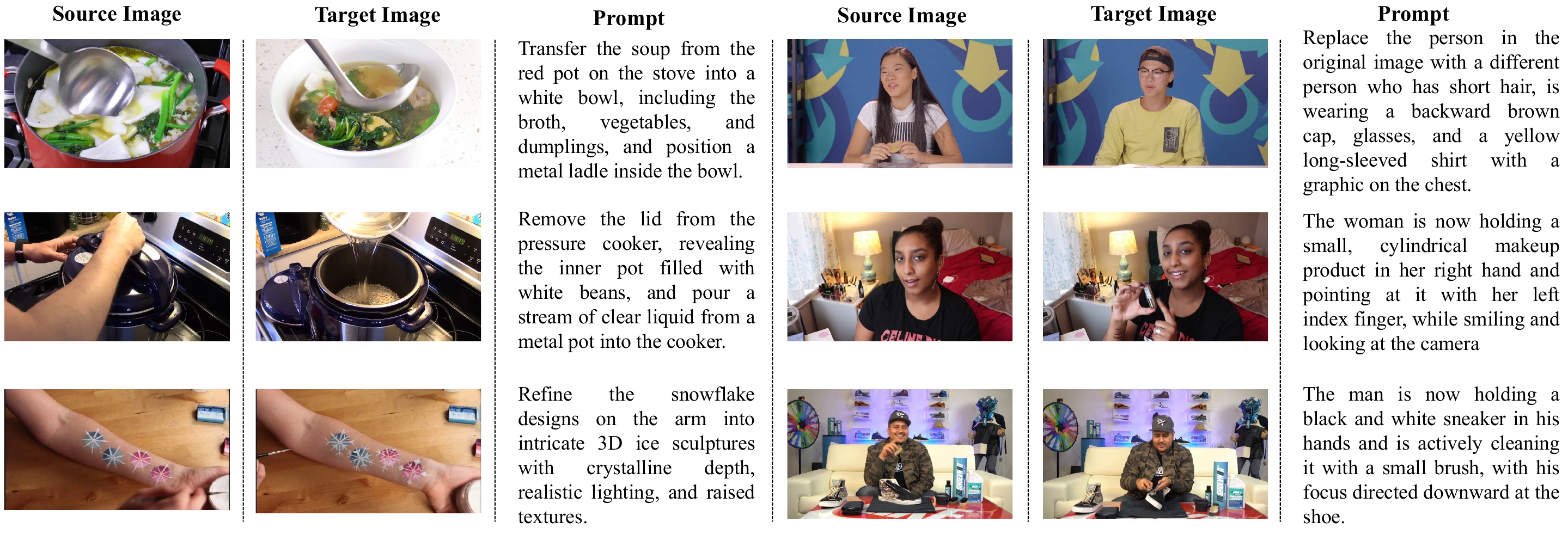}
    \caption{Examples of image pairs extracted from videos, along with their corresponding generated prompts.}
    
    \label{fig:image_pair_case}
\end{figure}

\textbf{Image-pair Data.}
\label{sec:image_pair_detail}
Similarly, a large-scale image manipulation dataset comprising nearly 30 million image pairs is constructed from tutorial videos~\cite{miech2019howto100m}. These videos capture naturally occurring and often complex visual transformations, providing diverse and realistic variations that help establish strong semantic alignment between image pairs. The construction pipeline is described below.
Key frames are sampled from over 300k videos, while low-motion or scaling-dominated frames are filtered based on inter-frame transformations, and blur detection is further applied to remove low-quality frames. For each video, image pairs are formed from the extracted frames, and pairwise similarities are computed using CLIP embeddings~\cite{radford2021learning}. Pairs with similarity scores within a predefined range, i.e., [0.75, 0.95], are retained to exclude both near-duplicate and semantically unrelated pairs. Finally, Qwen3-VL-30B-A3B-Instruct~\cite{bai2025qwen3} is used to generate textual prompts describing the visual differences between selected image pairs.
Figure~\ref{fig:image_pair_case} shows examples of the constructed image pairs. 

\textbf{Interleaved Image-text Data.}
\label{sec:interleave_data}
Inspired by prior work~\cite{bagel,cui2025emu3}, both web and video data are leveraged as key sources for constructing interleaved image-text data. For web data, following~\cite{bagel}, around 10 million interleaved samples are first built from OmniCorpus~\cite{li2024omnicorpus}. Beyond the basic filtering in~\cite{bagel}, Qwen3-32B~\cite{yang2025qwen3} is used to regenerate the textual content, improving fluency and coherence, and subject-aware question-answer pairs are further introduced for augmentation. For video data, up to 8 key frames are extracted from each video in the general T2V corpus, and Qwen3-VL-30B-A3B-Instruct~\cite{bai2025qwen3} is employed to generate textual transitions between frames, yielding 2 million additional video-derived samples.

\subsection{Diverse Image Editing and Image-to-Video Editing Data}
\label{sec:i2i_i2v_data}
\begin{figure}[t]
    \centering
    \includegraphics[width=\linewidth]{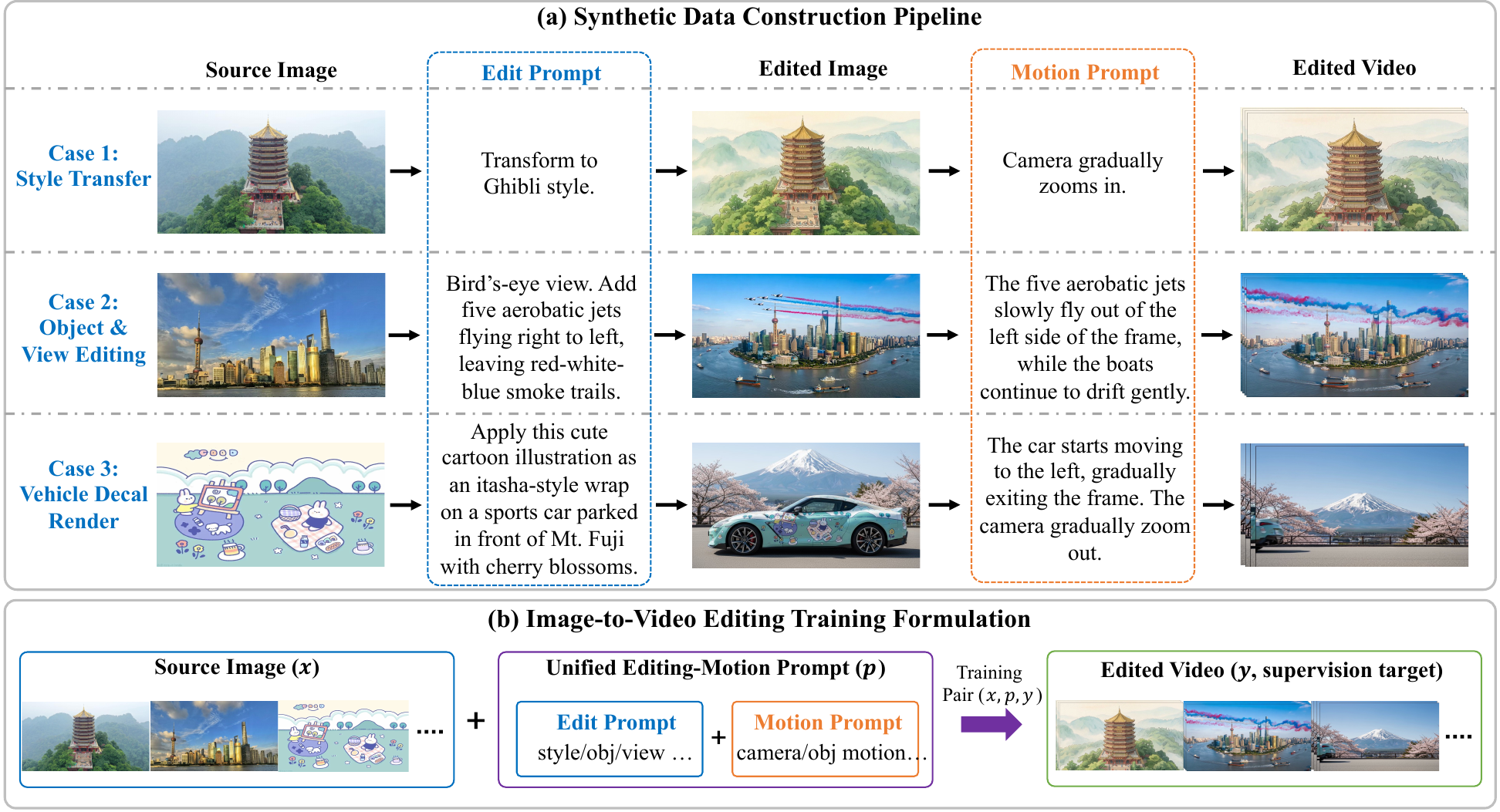}
    \caption{Example of generated Image Editing and Image-to-Video Editing Data.}
    \label{fig:i2i_i2v_case}
\end{figure}

Compared with image editing, constructing high-quality and diverse video-to-video editing data at scale remains significantly more difficult. Meanwhile, image-to-image editing has benefited from much more mature models and data resources. This suggests a practical route for improving video editing: reformulating part of the video editing problem as image-to-video editing, such that image-level editing capability can be transferred to video generation and eventually benefit video-to-video editing. In this way, image editing data serves not only as an auxiliary source of supervision, but also as a means to enrich the diversity and effectiveness of video editing training.
Diverse image editing prompts are constructed through two complementary mechanisms. The first starts from a large pool of real-world user instructions, from which multiple candidate prompts are sampled for each source image; an MLLM then selects the most suitable candidate and rewrites it into the final editing prompt. The second maintains a dynamic editing prompt bank to encourage diversity. Conditioned on the source image and the current prompt bank, the MLLM generates a new editing instruction with high semantic distinctiveness. Editing prompts with high novelty are inserted into the bank, while less diverse ones are discarded once the bank reaches its capacity. After edited images are obtained, corresponding motion prompts are further generated by the MLLM to synthesize target videos.
This process yields two types of training data: image editing triplets 
(Source Image, Edited Image, Edit Prompt), and image-to-video triplets 
(Source Image, Video, Edit Prompt+Motion Prompt). These data provide diverse supervision for transferring image editing knowledge to video editing. Examples can be found in Fig.~\ref{fig:i2i_i2v_case}.

\subsection{High-quality Video-to-Video Editing Data}
\label{sec:hq_v2v_data}
\begin{figure}[t]
\centering
\includegraphics[width=1.0\linewidth]{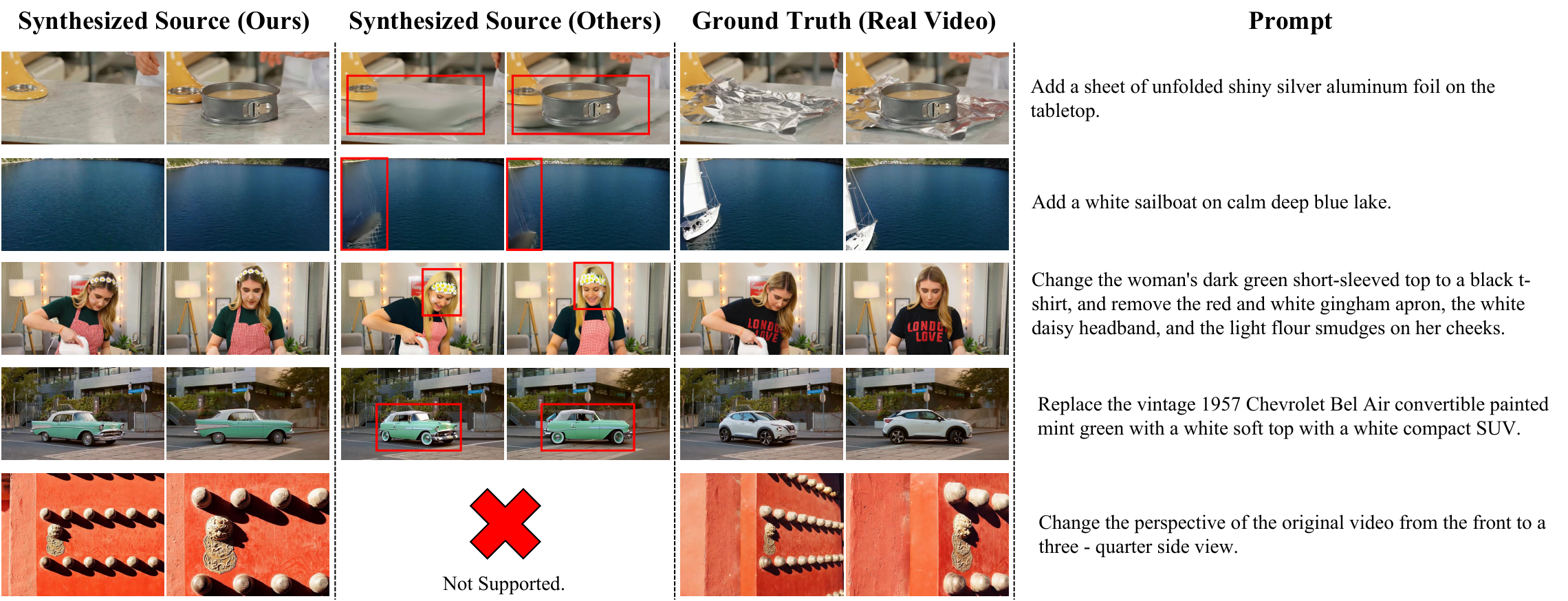}
\caption{Examples of generated propagation-based edit data. Compared with existing methods, our method yields higher-quality results. DiffuEraser~\cite{li2025diffueraser} (Rows 1-2) introduces obvious visual artifacts, while VACE~\cite{jiang2025vace} (Rows 3-4) suffers from character inconsistency and unnatural identical vehicle shape. Furthermore, our method enables more editing tasks (Row 5) that are unattainable by previous works.}
\label{fig:appendix_propagation_case}
\end{figure}
\textbf{Propagation-based Data Boosting.}
We first construct initial addition and removal data with DiffuEraser~\cite{li2025diffueraser} and replacement data with VACE~\cite{jiang2025vace}, but these data suffer from artifacts and limited edit diversity. For instance, the removal data contains visible artifacts, while the replacement samples are constrained to generating objects with shapes consistent with the originals, which can degrade model performance. To address these issues, we first train a base propagation model on the initial editing data mentioned above, where the model takes a source video, an edited first frame, and an editing prompt as input to generate the target edited video. We then combine this propagation model with a strong image editing model to build high-quality video editing data for common tasks such as addition, removal, replacement, and style transfer. Benefiting from the high quality of the edited first frames produced by the image editing model, the resulting edited videos also exhibit strong visual quality. To further improve the quality of data, we swap the source and edited video pairs and regenerate matching prompts with the MLLM for addition, removal and replacement tasks.
Examples can be found in Fig.~\ref{fig:appendix_propagation_case}.

\begin{figure}[t]
\centering
\includegraphics[width=0.99\linewidth]{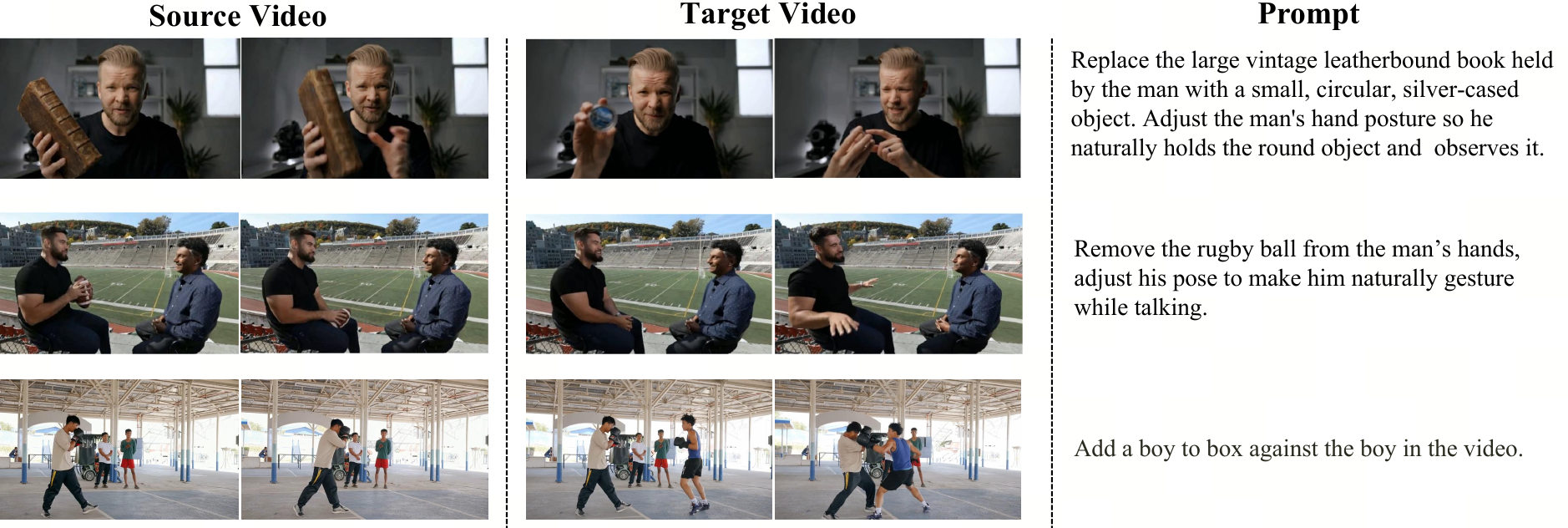}
\caption{Examples of generated motion-aware editing data. Our approach successfully synthesizes natural motions while maintaining consistency with the source videos. Specifically, it rationalizes human motions after object replacement and removal (Rows 1-2), and rationalizes interactive motions after adding a person (Row 3).
}
\label{fig:appendix_motion_aware}
\end{figure}
\textbf{Human Motion-aware Data.}
A strong video editing model should not only edit the target region accurately, but also adapt the surrounding scene to the consequences of the edit, especially in human--object interaction scenarios where object changes may naturally alter human pose and motion. However, such motion-aware editing data is difficult to collect and is poorly covered by existing synthesis pipelines. To address this, we propose a dual-branch data synthesis framework that combines the complementary strengths of image-to-video (I2V) generation and video-to-video (V2V) editing. Given an edited first frame and the source video, the I2V branch introduces motion adaptation, while the V2V branch preserves source motion consistency. Their outputs are fused with weighted guidance, enabling a controllable trade-off between motion preservation and action adaptation. 

Let $V$ denote the source video, $I$ the edited first frame, and $T_{I2V}$ and $T_{V2V}$ the prompts used in the two branches. We define
\begin{equation}
\begin{aligned}
\hat{\epsilon} = \;& \alpha \big( w^{I2V}_{Full} \cdot \epsilon(T_{I2V}, \varnothing, I) - w^{I2V}_{T} \cdot \epsilon(\varnothing, \varnothing, I) - w^{I2V}_{I} \cdot \epsilon(T_{I2V}, \varnothing, \varnothing) \big) \\
+\;& \beta \big( w^{V2V}_{Full} \cdot \epsilon(T_{V2V}, V, \varnothing) - w^{V2V}_{T} \cdot \epsilon(\varnothing, V, \varnothing) - w^{V2V}_{V} \cdot \epsilon(T_{V2V}, \varnothing, \varnothing) \big) \\
\\
\text{s.t.} \;& w^{I2V}_{Full} - w^{I2V}_{T} - w^{I2V}_{I} = 1, \quad w^{V2V}_{Full} - w^{V2V}_{T} - w^{V2V}_{V} = 1, \quad \alpha + \beta = 1
\end{aligned}
\end{equation}
Here, $w^{I2V}_{Full}$, $w^{I2V}_{T}$, and $w^{I2V}_{I}$ are the classifier-free guidance weights for the full, text-dropped, and image-dropped conditions in the I2V branch, respectively. $w^{V2V}_{Full}$, $w^{V2V}_{T}$, and $w^{V2V}_{V}$ are defined analogously for the V2V branch. The coefficients $\alpha$ and $\beta$ control the contributions of the two branches. This formulation allows the I2V branch to emphasize action adaptation while the V2V branch preserves source motion consistency. Examples of motion-aware editing data generated by this method are shown in Fig.~\ref{fig:appendix_motion_aware}.

\subsection{Reference-image-guided Video Generation and Editing Data}
\label{sec:r2v}
We construct training data for two tasks: reference-to-video (R2V) and reference+video-to-video (RV2V). R2V data spans two domains, general objects and persons, each requiring a tailored pipeline. RV2V data is then synthesized on top of R2V via instruction-based video editing.

\textbf{General-object R2V.}
For each source video, we sample keyframes and prompt an MLLM to identify the 3 to 5 most salient objects and, for each, author an editing instruction that extracts the object \emph{and re-places it into a different scene}. This explicit scene change mitigates the copy-paste shortcut caused by identical reference and target backgrounds. A high-quality keyframe and each instruction are then passed to an image editor to obtain one reference image per object. Finally, the MLLM produces an R2V caption from the (reference, keyframe) pair.

\textbf{Person R2V.}
Image editors often fail to preserve facial identity, so we avoid the editor for human references and instead exploit identity recurrences in long-form video. Clips are first grouped by their parent video or episode, and a face embedding is computed for every clip. For each high-quality target clip, we then retrieve a same-identity reference clip from the same video or episode, with cross-episode filtering to enforce visual diversity. The matched clip is finally downloaded and the person is cropped as a full-body reference image. Sourcing references from real footage rather than editor outputs guarantees identity preservation.

\textbf{Reference + video-to-video.}
RV2V requires triplets $(\text{input video}, \text{reference}, \text{target video})$ in which the input lacks the referenced object. Since such triplets rarely occur naturally, we synthesize them with a previously trained intermediate-version video editor. For each R2V sample, an MLLM authors an instruction that removes or replaces the referenced object in the target video, and the editor applies it to produce the RV2V input video. The original target and reference complete the triplet.

\subsection{Reference-video-guided Video Generation Data}
\label{sec:motion_transfer_data}
\begin{figure}[t]
\centering
\includegraphics[width=0.98\linewidth]{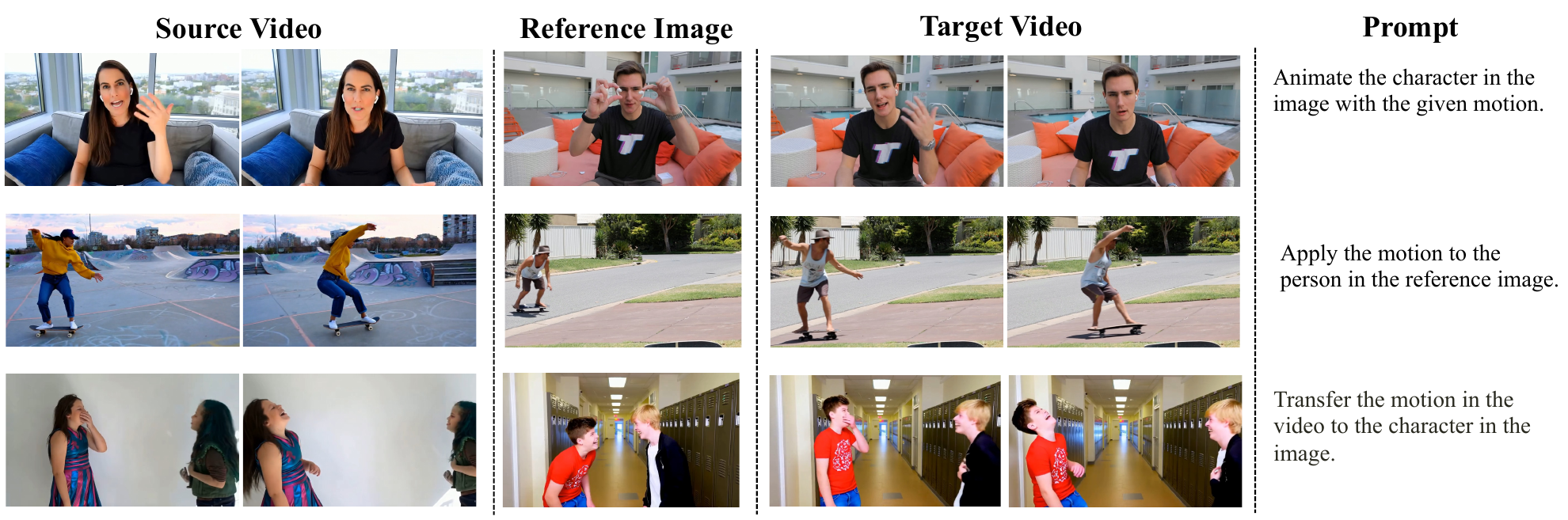}
\caption{Examples of generated motion-transfer editing data.}
\label{fig:motion_transfer_data}
\end{figure}
Beyond reference-image-guided video generation and editing, we further explore reference-video-guided settings. Compared with images, videos provide richer temporal information, and therefore we focus in particular on motion transfer.
Motion transfer refers to animating the person in an image using the motion of a person in a reference video. Training data for this task requires triplets of \mbox{$\langle$reference video, image, target video$\rangle$}. We first extract DWPose~\cite{yang2023effective} from real videos, and then use Bernini’s pose-to-video capability to generate a reference video with the same motion. The reference image is obtained by randomly sampling a frame from the target video. In this way, the triplets required for training can be constructed. Examples of motion transfer data are shown in Fig.~\ref{fig:motion_transfer_data}.

\subsection{Reasoning-augmented Video Data}
Inspired by recent unified generation methods~\cite{bagel, omnigen2, emu3}, we incorporate explicit Chain-of-Thought (CoT) reasoning into video editing with an MLLM planner. We consider both self-text reasoning, which rewrites editing instructions into structured intermediate prompts, and self-vision-text reasoning, which introduces visual intermediate states by decomposing video editing into image-level reasoning and video-level generation. This design improves editing fidelity and temporal coherence.

\begin{figure}[t]
\centering
\includegraphics[width=1\linewidth]{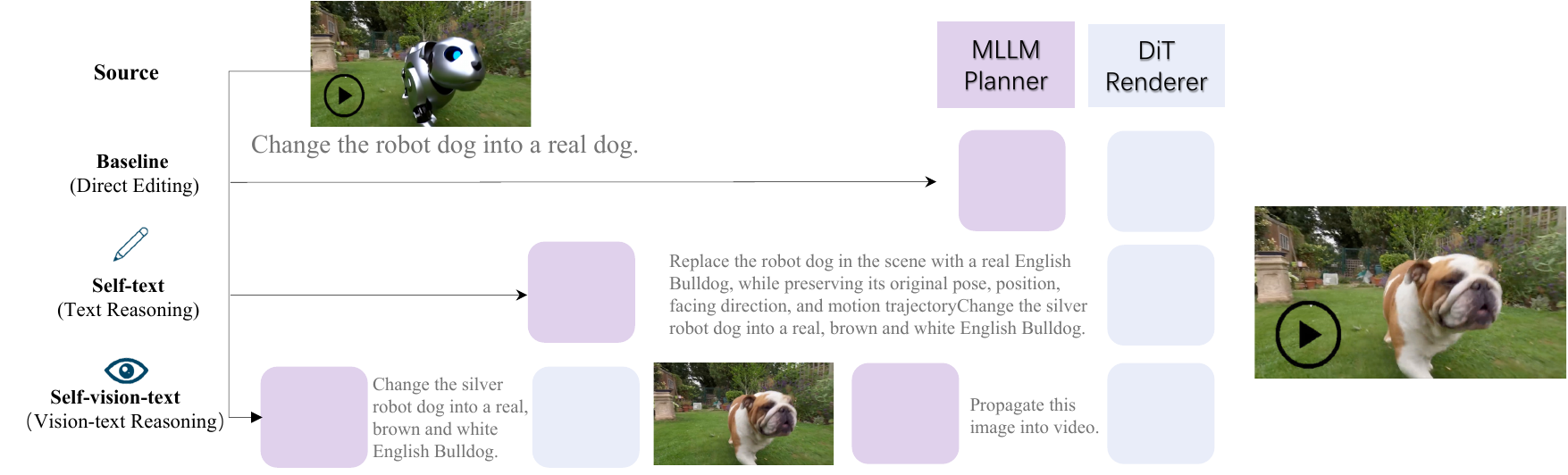}
\caption{Illustration of our reasoning pattern for reasoning-augmented video editing.}

\label{fig:cot_data_illustration}
\end{figure}

\textbf{Self-text Reasoning.}
Large-scale text-only CoT data is constructed to provide explicit reasoning supervision for video editing. The resulting dataset contains approximately 1M samples covering diverse editing tasks, including completion, addition, modification, and reasoning-driven transformations. To build this dataset, an MLLM is prompted with the source video, target video, and original editing instruction. The MLLM is then asked to rewrite the original prompt into a more detailed, structured, and semantically enriched editing instruction, which serves as the explicit reasoning signal.

\textbf{Self-vision-text Reasoning.}
While self-text reasoning provides explicit reasoning in the language space, it lacks direct grounding in visual transformations. To overcome this limitation, self-vision-text reasoning incorporates visual intermediate states into the reasoning process, decomposing video editing into two stages: image-level reasoning and video-level generation.
Given a source video and an editing instruction, the model first performs image editing on the initial frame, guided by textual reasoning, to produce an edited frame that reflects the intended transformation. This edited frame serves as a visual intermediate representation, grounding the reasoning process in the visual domain. 
Conditioned on this intermediate representation, the model then generates the target video by propagating the edits while preserving temporal consistency. 
This two-stage formulation bridges spatial reasoning and temporal generation, resulting in improved editing fidelity and temporal coherence.

As illustrated in Fig.~\ref{fig:cot_data_illustration}, our self-text reasoning refines and expands upon the initial editing instruction. Furthermore, our self-vision-text reasoning introduces an intermediate visual state to guide the video editing process, providing explicit visual grounding. Both approaches offer richer contextual information than the baseline method.

\begin{table*}[!t]
\centering
\fontsize{7}{8}\selectfont
\setlength{\tabcolsep}{6pt}
\renewcommand{\arraystretch}{1.1}
\caption{Statistics of key generation and editing training data used in the second phase of Stage II and the first phase of Stage IV.}

\begin{adjustbox}{max width=\textwidth}
\begin{tabular}{
>{\raggedright\arraybackslash}m{3.2cm}
>{\centering\arraybackslash}m{1.2cm}
>{\raggedright\arraybackslash}m{9.0cm}
}
\toprule
\textbf{Dataset} & \textbf{Weight} & \textbf{Information} \\
\midrule
\multicolumn{3}{l}{\textit{T2I — Text-to-Image Generation}} \\
\midrule
Inhouse T2I                         & 20.00 & Internal high-quality text-to-image dataset. \\
\midrule
\multicolumn{3}{l}{\textit{T2V — Text-to-Video Generation}} \\
\midrule
Inhouse T2V                         & 30.00 & Internal high-quality text-to-video dataset. \\
\midrule
\multicolumn{3}{l}{\textit{I2I — Image-to-Image Editing}} \\
\midrule
UniREdit-100K~\cite{han2025unireditbench}                 & 1.50  & Open-source unified editing dataset. \\
General-R2I                         & 2.80  & Constructed from the general r2v pipeline (Sec.~\ref{sec:r2v}); key frames as target. \\
Pico-Banana-400K~\cite{qian2025pico}  & 4.60  & Open-source Pico-Banana 400K single-SFT subset. \\
Diverse I2I                         & 5.00  & Diverse I2I data constructed following the I2I pipeline (Sec.~\ref{sec:i2i_i2v_data}). \\
Inhouse I2I                         & 26.10 & Internal instruction-based image editing dataset. \\
\midrule
\multicolumn{3}{l}{\textit{I2V — Image / Reference-to-Video Generation}} \\
\midrule
OpenS2V-Top200K~\cite{yuan2025opens2v}               & 0.05  & Open-source reference-to-video data. We selected 200K high-quality pairs from this, and applied affine transformations to the reference images as data augmentation. \\
Frame-to-Video                      & 0.15  & I2V data conditioned on first, first-last, or first-mid-last frames.  \\
Diverse I2V                         & 0.40  & Diverse I2V editing data constructed following the I2V pipeline (Sec.~\ref{sec:i2i_i2v_data}). \\
Person-R2V                          & 1.30  & Built via the Person R2V pipeline (Sec.~\ref{sec:r2v}). \\
General-R2V                         & 1.60  & Constructed via the General R2V pipeline (Sec.~\ref{sec:r2v}). \\
\midrule
\multicolumn{3}{l}{\textit{V2V — Video-to-Video Editing}} \\
\midrule
Video-Extension                     & 0.10  & Split video into two parts, take the second part as target video. \\
Video-Completion                    & 0.10  & Split video into three parts, take the middle part as target video. \\
Senorita-Controllable~\cite{zi2025senorita}         & 0.10  & Open-source controllable video editing data. \\
Sketch-to-Video                     & 0.10  & Sketch-conditioned video generation pairs. Sketch is detected with OpenCV Canny. \\
Inpainting-NoMask                   & 0.10  & Mask-free video inpainting pairs. We use GroundingDINO~\cite{liu2023grounding} and SAM2~\cite{ravi2024sam2segmentimages} to perform object segmentation. \\
Colorization                        & 0.10  & Video colorization pairs. \\
Movie-with-Subtitles                & 0.10  & Subtitle-removal pairs from movie clips. \\
Video2Mask                          & 0.10  & Video-to-mask paired data. We use GroundingDINO~\cite{liu2023grounding} and SAM2~\cite{ravi2024sam2segmentimages} to perform object segmentation.\\
Pose2Video                          & 0.15  & Pose-conditioned video generation pairs. Human skeleton detection is performed using DWPose~\cite{yang2023effective}.\\
SyncamVideo~\cite{bai2024syncammaster}   & 0.15  & Open-source SynCamVideo-Dataset video data. \\
TrajectoryCrafter~\cite{yu2025trajectorycrafter}  & 0.15  & Constructed via the Open-source TrajectoryCrafter model. \\
CameraClone~\cite{luo2025camclonemaster}   & 0.15  & Open-source CamCloneMaster video data. \\
VACE-HQ~\cite{jiang2025vace}                   & 0.30  & VACE-generated data with human filtering (Sec.~\ref{sec:hq_v2v_data}). \\
Motion-aware Editing                & 0.60  & Constructed via the motion-aware data pipeline (Sec.~\ref{sec:hq_v2v_data}). \\
Propagation-based Editing           & 1.00  & Constructed via the propagation-based data pipeline (Sec.~\ref{sec:hq_v2v_data}). \\
\midrule
\multicolumn{3}{l}{\textit{IV2V — Reference-guided Video Editing}} \\
\midrule
Motion-Transfer                     & 0.10  & Constructed via the motion-transfer data pipeline (Sec.~\ref{sec:motion_transfer_data}). \\
Propagation                         & 0.40  & Propagation data built from V2V first-frame extraction. \\
Motion-aware Editing Ref            & 0.60  & Constructed from motion-aware editing data by using an image editing model to extract the edited object from the first frame as the reference image. \\
Propagation-based Editing Ref       & 1.05  & Constructed from Propagation-based Editing data by using an image editing model to extract the edited object from the first frame as the reference image. \\
Person-RV2V                         & 1.05  & Person-centric RV2V replacement data. \\
\bottomrule
\end{tabular}
\end{adjustbox}
\label{tab:training_data}
\end{table*}
\section{Training and Inference}

\subsection{Training Pipelines}
\begin{table*}[t]
    \centering
    \caption{Training settings across different stages, where Res. denotes resolution, V.P. denotes video pairs, I.P. denotes image pairs, Int. denotes interleaved image-text data, Und. denotes understanding data, and CoT denotes reasoning-augmented video data.}
    \label{tab:training_strategy}
    \scalebox{0.85}{
    \begin{tabular}{c|c|c|c|c|c|c|c|c|c|c|c|c|c|c|c} 
    \toprule
    \textbf{Stage} & \textbf{Optimized}
    & \textbf{Res.}
    & \textbf{LR}
    & \textbf{EMA}
    & \textbf{T2I}
    & \textbf{T2V}
    & \textbf{I2I}
    & \textbf{V2V}
    & \textbf{I2V}
    & \textbf{IV2V}
    & \textbf{V.P.}
    & \textbf{I.P.}
    & \textbf{Int.} 
    & \textbf{Und.}
    & \textbf{CoT} \\
    \midrule
    \multirow{1}{*}{I}
        & MLLM & 256p & 1e-5 & 0.999  & 13\%  & 19\%  & 3\%   & 1\%   & 1\%   & 1\%   & 15\%  & 21\%  & 6\%   & 20\% & -- \\
    \midrule
    \multirow{2}{*}{II}
        & DiT          & 480p & 1e-5 & 0.9995 & 31\%  & 42\%  & 4\%  & 0.4\%  & 0.4\%  & 0.3\%  & 11\%  & 11\%  & --    & -- & -- \\
        & DiT          & 480p & 1e-5 & 0.9999 & 20\%  & 30\%  & 40\% & 3.3\%  & 3.5\%  & 3.2\%  & --    & --    & --    & -- & -- \\
    \midrule
    \multirow{2}{*}{III}
        & Connector    & 256p & 1e-5 & --  & 80\% & 20\% & -- & -- & -- & -- & -- & -- & -- & -- & -- \\
        & Connector    & 256p & 1e-5 & --  & 50\% & 20\% & 20\% & 10\% & -- & -- & -- & -- & -- & -- & -- \\
    \midrule
    \multirow{2}{*}{IV}
        & All          & 480p & 1e-5 & 0.9995 & 16\%  & 24\%  & 32\% & 2.6\%  & 2.8\%  & 2.6\%  & --    & --    & --    & 20\% & -- \\
        & All          & 480p & 1e-5 & 0.999 & 12\%  & 18\%  & 24\% & 2\%  & 2\%  & 2\%  & --    & --    & --    & 20\% & 20\% \\
    \bottomrule
    \end{tabular}
    }
\end{table*}

To fully exploit the understanding capability of the MLLM and the synthesis capability of the diffusion model, we adopt a four-stage training pipeline as shown in Table~\ref{tab:training_strategy}. 
The composition of the key generation and editing data is summarized in Table~\ref{tab:training_data}.
We first train the MLLM-based planner and the DiT-based renderer separately, then train the MLP connector to bridge the two components, and finally lightly co-train the full model to align semantic planning with visual rendering.
This design preserves the strengths of both components while avoiding excessive interference during early training.

\textbf{Stage I: MLLM pretraining.}
In Stage I, we train the MLLM planner together with the ViT embedding decoder to predict target visual semantics in the ViT embedding space. Training is conducted with the joint objective $\lambda_{\mathrm{text}} \mathcal{L}_{\mathrm{ntp}} + \lambda_{\mathrm{visual}} \mathcal{L}_{\mathrm{visual}}$, where $\lambda_{\mathrm{text}}=0.2$ and $\lambda_{\mathrm{visual}}=1$. 
The goal of this stage is to transform the MLLM from a pure understanding model into a semantic planner that can infer target visual representations from multimodal context.

Training follows a progressive data curriculum. Large-scale text-to-image data is used first to establish image generation ability in the semantic space. The training corpus is then expanded to include text-to-video, image-pair, and video-pair data, enabling the planner to model not only image and video generation, but also image and video editing within a unified semantic space. To preserve the pretrained language and multimodal reasoning capabilities of the MLLM, multimodal understanding data and text understanding data are further incorporated at this stage. Training is performed at 256p resolution and 2 fps.

To improve robustness across heterogeneous tasks, we adopt a task-dependent mask ratio strategy. Specifically, the mask ratio $r \in [0,1]$ is randomly sampled from a Beta distribution,
\begin{equation}
r \sim \mathrm{Beta}(\alpha, \beta),
\end{equation}
where $(\alpha, \beta)$ are specified for each task. This design provides a flexible way to control the amount of visible target information under different training objectives. As the task input becomes more informative, e.g., from text-only generation to image/video-conditioned editing, we gradually increase $\alpha$ and decrease $\beta$, which shifts the distribution of $r$ toward probability $1.0$. As a result, a larger portion of target visual tokens is masked during training, reducing information leakage from the visible target tokens and forcing the planner to infer the masked semantics from higher-level multimodal context. This strategy is particularly important for editing tasks, where the source input is highly correlated with the target and may otherwise make semantic prediction overly easy. The detailed configuration is summarized in Table~\ref{tab:mask_ratio_config}.

Overall, Stage I equips the planner with broad semantic prediction ability across heterogeneous generation, editing, and understanding tasks, while the task-dependent mask ratio further improves its robustness under diverse semantic completion difficulties.

\begin{table}[t]
\centering
\caption{Mask ratio configuration during MLLM planner training. The mask ratio is sampled from a task-dependent Beta distribution $\mathrm{Beta}(\alpha,\beta)$.}
\small
\label{tab:mask_ratio_config}
\begin{tabular}{ccccccc}
\toprule
\textbf{Parameter} & \textbf{T2I} & \textbf{T2V} & \textbf{I2I} & \textbf{I2V} & \textbf{V2V} & \textbf{IV2V} \\
\midrule
$\alpha$ & 5.0  & 8.0  & 8.0  & 10.0 & 12.0 & 12.0 \\
$\beta$  & 1.1  & 1.05 & 1.05 & 1.0  & 0.9  & 0.9  \\
\bottomrule
\end{tabular}
\end{table}

\textbf{Stage II: DiT pretraining.}
In Stage II, we train the DiT-based renderer, together with its lightweight text encoder, e.g., T5, to endow it with strong generation and editing ability before coupling it with the MLLM planner. The renderer is optimized with $\mathcal{L}_{\mathrm{dit}}$ and conditioned on text features and source VAE features, allowing it to learn both high-fidelity synthesis and source-preserving editing.

In this stage, the renderer is trained on a large mixture of text-to-image, text-to-video, editing, image-pair, and video-pair data. Over 60K training steps,
the renderer processes approximately 20.3M sampled training instances in
total. While pair data is particularly beneficial for improving generalization and editing quality, it may also lead to weaker instruction following and inconsistencies in non-edited regions. We therefore adopt a linearly decayed sampling strategy for pair data, using a high ratio at the beginning of training and gradually reducing it to zero, such that the later stage relies on high-quality editing data to refine editing performance. Training in this stage is performed at 480p and 16 fps.%

To accommodate the varying optimization dynamics across these distinct tasks, we assign customized shift parameters and noise weighting schemes for each individual task, as summarized in Table~\ref{tab:noise_scheduler_config}. Following SD3~\cite{sd3} and Waver~\cite{zhang2025waver}, the logit-normal and mode functions are used for timestep sampling, as shown in Eq.~\ref{eq:logit_normal} and Eq.~\ref{eq:mode}, respectively. We adopt Lognorm(0.5, 1) for the image related tasks, Mode(1.29) for the video related tasks.

\begin{table}[t]
\centering
\caption{Training noise scheduler configuration.}
\label{tab:noise_scheduler_config}
\small
\begin{tabular}{lcccccc}
\toprule
\textbf{Parameter} & \textbf{T2I} & \textbf{I2I} & \textbf{T2V} & \textbf{I2V} & \textbf{V2V} & \textbf{IV2V} \\
\midrule
Weighting & logit-normal & logit-normal & mode & mode & mode & mode \\
Shift     & 3.0 & 4.0 & 3.0 & 5.0 & 5.0 & 5.0 \\
\bottomrule
\end{tabular}
\end{table}

\begin{equation}
\label{eq:logit_normal}
\pi_{\text{ln}}(t; m, s) = \frac{1}{s\sqrt{2\pi}} \frac{1}{t(1-t)} \exp\left( -\frac{(\text{logit}(t) - m)^2}{2s^2} \right),
\end{equation}

\begin{equation}
\label{eq:mode}
f_{\text{mode}}(u; s) = 1 - u - s \cdot \left( \cos^2\left( \frac{\pi}{2} u \right) - 1 + u \right).
\end{equation}

\textbf{Stage III: Connector training.}
In Stage III, we freeze the MLLM planner and the DiT renderer and train only the MLP connector to map the MLLM hidden states into the conditioning representation required by the renderer. 
Training follows a two-phase curriculum at 256p. 
We first use text-to-image and text-to-video data to establish the semantic-to-visual alignment for generation. 
Then, image-to-image and video-to-video editing data are incorporated to extend this alignment to editing tasks. 
This stage provides a stable interface between the pretrained planner and renderer before joint training.

\textbf{Stage IV: Joint training.}
In Stage IV, we jointly train the MLLM planner, MLP connector, and DiT renderer to further align semantic planning with visual rendering. 
The model is optimized using the objective in Eq.~\ref{eq:total_loss}, with $\lambda_{text}=0.2$ and $\lambda_{visual}=\lambda_{dit}=1$. 
Specifically, during training, the text, source ViT tokens, and masked target ViT tokens are fed into the MLLM. The MLLM hidden states corresponding to the text, source ViT tokens, and unmasked target ViT tokens are extracted as the conditioning input for the diffusion model. Meanwhile, the hidden states corresponding to the masked target ViT tokens are fed into the ViT embedding decoder, where 
$\mathcal{L}_{visual}$ is computed.

Joint training is conducted at 480p and 16 fps on a mixture of high-quality image and video generation/editing data together with text-only and multimodal understanding data. The understanding data helps preserve the MLLM's language and multimodal reasoning capabilities, while the generation and editing data encourage the emergence of a stable semantic-to-visual interface between the planner and the renderer. In the later phase of joint training, we additionally introduce reasoning-augmented Chain-of-Thought (CoT) data to enhance structured reasoning for video editing. This encourages the model to perform more explicit semantic planning over object dynamics, temporal transitions, and editing intent before rendering the final output.

Compared with the preceding stages, Stage IV uses only light co-training for a relatively small number of steps. This is sufficient to align the planner and renderer while preserving the pretrained strengths of both. As a result, the MLLM retains strong understanding and reasoning ability, the renderer preserves high-fidelity generation and editing performance, and the overall system learns to translate multimodal reasoning into faithful visual outputs.

\subsection{Inference Strategy}

\textbf{ViT embedding planning via MLLM.}
During inference, the target visual tokens are initialized as masked, and the MLLM progressively predicts the target semantic tokens following standard masked generative inference~\cite{chang2022maskgit,li2024autoregressive}. Unless otherwise specified, we use 25 iterative planning steps to predict the full target semantic embedding sequence. At each planning step, the predicted semantic features are decoded into target ViT embeddings by the ViT embedding decoder via flow matching, where the decoder performs 5 diffusion denoising steps in the ViT embedding space. For this diffusion refinement stage, the text and image guidance scales are set to 1.2 and 1.0, respectively.

After all target ViT embeddings are obtained, they are fed back into the MLLM together with the textual embedding and source ViT embeddings to produce contextualized hidden states for conditioning the DiT-based renderer. In practice, the iterative inference of the MLLM-based planner introduces only negligible overhead compared with the subsequent DiT rendering stage, with runtime that is almost negligible relative to the DiT renderer. This indicates that the additional cost of multi-step semantic planning is negligible relative to the overall diffusion sampling cost, while still providing high-quality semantic guidance for downstream rendering.

\textbf{Visual target rendering via DiT.}
The DiT-based renderer performs latent-space denoising under multi-source guidance from source video VAE features, source image VAE features, text features, and target semantic embeddings. The flow shift is set to 5.0 for the DiT-based renderer. The renderer performs 50 denoising steps for text-to-video generation and 40 denoising steps for reference-to-video generation, video-to-video editing, and reference-guided video-to-video editing.

To control the contribution of different conditions, we decompose the final prediction into an unconditional base term and four incremental guidance terms associated with the source video VAE features, source image VAE features, text features, and target semantic embeddings, respectively. Specifically, let $\epsilon_{\varnothing,\varnothing,\varnothing,\varnothing}$ denote the prediction without any condition, $\epsilon_{\varnothing,\varnothing,\mathrm{vid},\varnothing}$ the prediction conditioned only on the source video VAE features, $\epsilon_{\varnothing,\varnothing,\mathrm{vid},\mathrm{img}}$ the prediction conditioned on both source video and source image VAE features, $\epsilon_{\mathrm{txt},\varnothing,\mathrm{vid},\mathrm{img}}$ the prediction additionally conditioned on text features, and $\epsilon_{\mathrm{txt},\mathrm{tgt},\mathrm{vid},\mathrm{img}}$ the prediction further conditioned on target semantic embeddings. The incremental contributions are defined as
\begin{align}
\Delta_{\mathrm{vid}}
&= \epsilon_{\varnothing,\varnothing,\mathrm{vid},\varnothing}
- \epsilon_{\varnothing,\varnothing,\varnothing,\varnothing}, \\
\Delta_{\mathrm{img}}
&= \epsilon_{\varnothing,\varnothing,\mathrm{vid},\mathrm{img}}
- \epsilon_{\varnothing,\varnothing,\mathrm{vid},\varnothing}, \\
\Delta_{\mathrm{txt}}
&= \epsilon_{\mathrm{txt},\varnothing,\mathrm{vid},\mathrm{img}}
- \epsilon_{\varnothing,\varnothing,\mathrm{vid},\mathrm{img}}, \\
\Delta_{\mathrm{tgt}}
&= \epsilon_{\mathrm{txt},\mathrm{tgt},\mathrm{vid},\mathrm{img}}
- \epsilon_{\mathrm{txt},\varnothing,\mathrm{vid},\mathrm{img}}.
\end{align}
Accordingly, the final prediction is
\begin{align}
\hat{\epsilon}
= \epsilon_{\varnothing,\varnothing,\varnothing,\varnothing}
+ \omega_{\mathrm{vid}} \Delta_{\mathrm{vid}}
+ \omega_{\mathrm{img}} \Delta_{\mathrm{img}}
+ \omega_{\mathrm{txt}} \Delta_{\mathrm{txt}}
+ \omega_{\mathrm{tgt}} \Delta_{\mathrm{tgt}},
\end{align}

where $\omega_{\mathrm{vid}}$, $\omega_{\mathrm{img}}$, $\omega_{\mathrm{txt}}$, and $\omega_{\mathrm{tgt}}$ are the corresponding guidance scales for source video VAE features, source image VAE features, text features, and target semantic embeddings, respectively. We further apply adaptive projected guidance~\cite{sadat2024apg} to reduce oversaturation.

The guidance scales used by the DiT-based renderer for different tasks are summarized in Table~\ref{tab:inference_guidance}. For text-to-video generation, where no source video is provided, the source video guidance term is not used.

\begin{table}[t]
\caption{Inference guidance scales for different tasks.}
\centering
\small
\begin{tabular}{lccccc}
\toprule
\textbf{Task} & \textbf{Steps} & $\omega_{\mathrm{txt}}$ & $\omega_{\mathrm{vid}}$ & $\omega_{\mathrm{img}}$ & $\omega_{\mathrm{tgt}}$ \\
\midrule
T2V  & 50 & 4.0 & --   & 1.0  & 0.5 \\
R2V  & 40 & 4.0 & 1.25 & 4.5  & 1.5 \\
V2V  & 40 & 4.0 & 1.25 & 1.25 & 0.5 \\
RV2V & 40 & 4.0 & 0.75 & 3.0  & 1.5 \\
\bottomrule
\end{tabular}
\label{tab:inference_guidance}
\end{table}

\section{Infrastructure}
\label{appendix:infrastructure}
\subsection{Training Infrastructure}

Training long-context video editing models with co-trained DiT and MLLM components posed substantial systems challenges in memory, computation, parallelism, and data loading. On the memory side, we optimized FSDP configurations and restructured the input pipeline to use direct index-scattering into pre-allocated buffers, reducing per-GPU memory from 72\,GB to 40\,GB. Combined with a custom activation offloading pipeline with pinned CPU memory pools and delayed-queue prefetch, these achieved a \emph{4.4$\times$} improvement for training sequence length. On the computation side, kernel-level optimizations including FlashAttention-4~\cite{zadouri2026flashattention4}, asynchronous QKV communication, TND memory layout preservation, and a high-performance RMSNorm kernel~\cite{quack2025} collectively yielded up to \emph{46\%} speedup. For parallelism, we adopted Ulysses-style sequence parallelism~\cite{jacobs2023deepspeed} for both DiT and MLLM, extending it to selectively unfrozen MLLM in joint training. We further implemented sequence packing with token-bucket batching and greedy bin-packing data loading to handle heterogeneous sequence lengths, together improving end-to-end throughput by $\sim$\emph{4.5$\times$}.

\textbf{Memory Optimization.} Video editing training involved extremely long sequences that imposed severe GPU memory pressure. We systematically profiled and optimized FSDP configurations, reducing per-GPU memory from 72\,GB to 40\,GB. Beyond FSDP tuning, we restructured the input preparation pipeline: instead of first concatenating all visual and textual tokens and then scattering them to target positions for sequence parallel, we directly index-scattered tokens into pre-allocated buffers, eliminating 17\,GB of intermediate memory allocation. For Stage IV, we implemented a custom activation offloading pipeline with a pinned CPU memory pool and a delayed-queue prefetch mechanism, overlapping D2H/H2D transfers with computation. Combined with padding and normalization optimizations, these strategies enabled stable training with 440K-token sequences---a \emph{4.4$\times$} improvement over the previous 100K-token limit.

\textbf{High-Performance Operators and Pipelines.}
We performed systematic, kernel-level optimizations tailored for our target GPU architecture. Key optimizations included applying FlashAttention-4~\cite{zadouri2026flashattention4} in the DiT and FlexAttention~\cite{dong2024flex} in the MLLM, implementing an asynchronous QKV communication pipeline, and eliminating redundant cross-attention communication. 
Furthermore, we maintained the TND memory layout to avoid costly transposes and placed \texttt{cu\_seqlens} tensors on the CPU to reduce device memory pressure. 
We also adopted the high-performance RMSNorm kernel from QuACK~\cite{quack2025}, which yielded an additional 5--10\% end-to-end training speedup. 
A unified attention backend ensures seamless deployment across heterogeneous GPU clusters.

\textbf{Parallelism Strategy.}
We employed FSDP for memory-efficient weight sharding combined with Ulysses-style sequence parallelism~\cite{jacobs2023deepspeed} for both DiT and MLLM components. For the DiT-based Renderer, sequence parallelism sharded tokens across GPUs along the sequence and head dimensions, enabling the processing of long video sequences. For the MLLM-based Planner, we extended Ulysses SP, achieving $2\times$ throughput at SP degree~4. SP was enabled only for long-sequence tasks to avoid unnecessary communication overhead on shorter inputs.

\textbf{Sequence Packing and Batch Forward.}
To improve GPU utilization under heterogeneous sequence lengths, we implemented a comprehensive sequence packing pipeline. Training samples were first sorted by sequence length for each sequence parallel group, achieving a $2\times$ throughput speedup. We then introduced batch forward for both MLLM and diffusion components: MLLM inputs were batched with FlashAttention variable-length kernels, while diffusion inputs were concatenated and processed jointly. To avoid cross-rank deadlocks caused by varying local batch sizes, we applied dummy-forward padding to ensure consistent execution across all ranks. We further introduced token-bucket batching, which grouped samples into discrete length buckets and applied per-bucket loss re-weighting to eliminate padding waste while preserving training dynamics. Together, these optimizations improved end-to-end throughput by $\sim$4.5$\times$.

\textbf{Dataloader Balance.}
Large-scale video editing training involved highly heterogeneous data,
including varying video lengths, resolutions, and editing operations,
which introduced significant computational imbalance across GPUs. We implemented a load-balanced data loader using greedy bin-packing to redistribute workloads across nodes at each iteration, achieving a max/min workload ratio below 1.01 and approximately 15\% throughput improvement.

\subsection{Inference Parallelism}
We adopted multi-GPU inference to further reduce latency. For the DiT model, we integrated DeepSpeed Ulysses~\cite{jacobs2023deepspeed} with asynchronous all-to-all communication for the QKV tensor. For the VAE module, we employed context parallelism along the temporal dimension with asynchronous conv cache transmission. Together, these achieved a speedup of over \emph{7.2$\times$}.

\subsection{Model Distillation}

To reduce the sampling cost of our diffusion model while preserving generation quality, we adopted a two-stage distillation strategy. In the first stage, we performed \emph{CFG distillation}~\cite{meng2023distillation}, which trained a student model to directly predict the CFG-combined output in a single forward pass, eliminating the need for dual (conditional and unconditional) evaluations at each sampling step and halving the per-step compute. In the second stage, we applied ReFlow \cite{liu2022reflow}, which straightened the learned probability flow ODE trajectories, enabling accurate generation with significantly fewer integration steps. By progressively reducing both per-step cost and the total number of required steps, this two-stage pipeline achieved substantial end-to-end inference speedup with minimal quality degradation. Finally, the distilled student model with only 4 NFEs achieved comparable quality to the teacher model with 80 NFEs.

\section{Experiments}
We evaluate Bernini on two complementary task families that together cover its capabilities as a unified framework: \emph{video editing} and \emph{video generation}. To enable a more comprehensive evaluation of video editing capabilities, we introduce Bernini-Bench (Sec.~\ref{app:bernini_bench}). Then we present main results on video editing (Sec.~\ref{sec:video_editing}) and analyze the contribution of reasoning-augmented editing (Sec.~\ref{sec:editing_with_reasoning}), followed by results on video generation (Sec.~\ref{sec:video_generation}). We also conduct ablation studies (Sec.~\ref{sec:ablation}) and discuss the generalizability of Bernini (Sec.~\ref{sec:generalization}).

\subsection{Implementation Details}

Qwen2.5-VL-7B~\cite{qwen2.5vl} is adopted as the MLLM-based planner, and Wan2.2-A14B~\cite{wan} serves as the DiT-based renderer. 
To better align with the pretrained conditioning distribution of Wan2.2, we retain the original T5 features in the textual condition of the DiT renderer. Specifically, only the penultimate-layer hidden states of the MLLM are passed through a lightweight zero-initialized one-layer MLP, and the projected features are then concatenated with the T5 features to form the final conditioning input. This design preserves Wan2.2's pretrained text-conditioning prior while introducing higher-level semantic guidance from the MLLM.
Unless otherwise specified, following prior work~\cite{team2025kling}, we enhance the user instruction with an additional multimodal large language model and feed the rewritten instruction into Bernini to further improve performance.

\subsection{Bernini-Bench}
\label{app:bernini_bench}

\begin{figure}[t]
    \centering
    \includegraphics[width=0.85\linewidth]{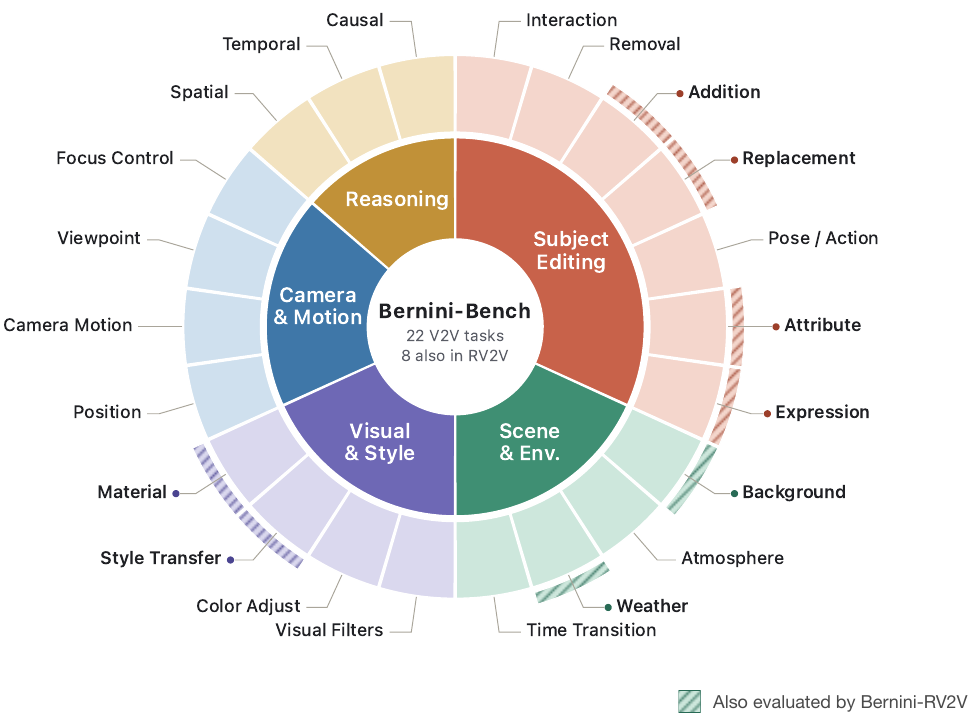}
    \caption{Overview of Bernini-Bench. Our benchmark spans 22 fine-grained V2V editing tasks across five dimensions: Subject Editing, Scene \& Environment, Visual \& Style, Camera \& Motion, and Reasoning. Hatched segments denote the 8 tasks also evaluated under the reference-image-guided video editing (Bernini-RV2V) setting.}
    \label{fig:bernini-bench}
\end{figure}

\textbf{Benchmark Construction.}
Currently, prevalent video editing benchmarks, such as OpenVE-Bench~\cite{he2025openve} and EditVerse~\cite{ju2025editverse}, predominantly focus on video-to-video editing, neglecting the reference+video-to-video paradigm. Moreover, these benchmarks are relatively limited in both the diversity of editing types and the variety of video content.
To provide a more comprehensive evaluation of video editing models, we manually build Bernini-Bench, a new benchmark for assessing editing performance across \emph{diverse task types} and \emph{real-world scenarios}. Bernini-Bench covers two input settings, \textit{text-guided video-to-video editing (V2V)} and \textit{reference-image-guided video editing (RV2V)}. It comprises \emph{300} test cases spanning \emph{22} editing categories, including action editing, position editing, edits involving causal reasoning, and edits with changes in camera focus, which are editing types not covered by other benchmarks. For each editing category, 10 cases were carefully selected, each accompanied by rich editing instructions (e.g., a wide range of target styles for style transfer). To better reflect real-world applications, we collected source videos from several free and open-source stock media platforms. The selected videos cover diverse editing-relevant attributes, including variations in human composition, shot scale, scene environment, camera motion, and visual complexity, and include both horizontal and vertical aspect ratios. The detailed statistics of Bernini-Bench are presented in Fig.~\ref{fig:bernini-bench}.

\textbf{Evaluation Metrics.}
Similarly to existing video editing benchmarks~\cite{ju2025editverse, he2025openve}, we evaluate the model performance across five dimensions: instruction following, source video consistency, reference image consistency, generation quality and overall score.
For a comprehensive performance evaluation, all dimensions (excluding overall performance) are set to be as orthogonal as possible for independent assessment. The specific criteria are as follows:
\begin{itemize}[leftmargin=*,itemsep=1pt,topsep=2pt]
    \item \textbf{Instruction Following (IF)}: Evaluates the model’s ability to accurately and faithfully execute textual editing instructions, such as correctly identifying the editing target and operation type.
    \item \textbf{Video Consistency (VC)}: Measures whether the non-edited regions of the video remain consistent before and after editing.
    \item \textbf{Reference Image Consistency (IC)}: Assesses the consistency of visual features (shape, color, texture, style) between the editing result and the given reference image. This metric is evaluated only for the RV2V task.
    \item \textbf{Generation Quality (GQ)}: Focuses on the video’s physical realism, edited content naturalness, as well as the presence of severe AI artifacts and obvious visual distortion.
    \item \textbf{Overall Score (OS)}: Evaluates whether the editing result meets the user’s expectations.
\end{itemize}

For actual evaluation, we adopt two approaches for each dimension: MLLM-based scoring, and human Side-by-Side (SBS) comparison.
For MLLM-based scoring, the model assigns a score ranging from 1 to 5 for each evaluation dimension. 
Specifically, samples where the model fails to respond to the instruction at all are excluded from the final score calculation for source video consistency, reference image consistency, and generation quality.
Since current MLLMs are unable to accurately judge issues such as small-scale distortions or unnatural artifacts when assessing generation quality, the corresponding results should be treated as reference only. We use GPT-5.4-2026-03-05 for evaluation.
The detailed prompts used for MLLM scoring can be found in the Appendix~\ref{app:eval_prompts}.

\subsection{Video Editing}
\label{sec:video_editing}

\begin{table}[t]
    \centering
    \small
    \caption{Quantitative results on Bernini-V2V and RV2V.}
    \label{tab:bernini_ben_res}
    \setlength{\tabcolsep}{3.5pt}
    \begin{tabular}{l|cccc|ccccc}
        \toprule
        \multirow{2}{*}{Method} & \multicolumn{4}{c|}{Bernini-V2V} & \multicolumn{5}{c}{Bernini-RV2V} \\
        \cmidrule(lr){2-5} \cmidrule(lr){6-10}
        & OS & IF & VC & GQ & OS & IF & VC & IC & GQ \\
        \midrule
        UniVideo~\cite{univideo}  & 2.44 & 2.58 & 3.30 & 3.16 & 2.36 & 2.67 & 3.15 & 2.87 & 2.82 \\
        VINO~\cite{vino2026}      & 2.85 & 3.08 & 3.14 & 3.26 & 2.25 & 2.64 & 2.17 & 3.51 & 3.06 \\
        Kling O3~\cite{team2025kling} & 3.05 & 3.25 & 3.09 & 3.44 & 3.14 & 3.41 & 3.14 & 3.61 & \underline{3.30} \\
        Wan2.7~\cite{wan}         & \underline{3.30} & \underline{3.57} & \underline{3.11} & \textbf{3.56} & \textbf{3.58} & \underline{3.82} & \textbf{3.48} & \underline{3.62} & \textbf{3.43} \\
        \midrule
        Bernini          & \textbf{3.49} & \textbf{3.60} & \textbf{3.65} & \underline{3.52} & \underline{3.55} & \textbf{3.86} & \underline{3.24} & \textbf{3.67} & \underline{3.30} \\
        \bottomrule
    \end{tabular}
\end{table}

\textbf{MLLM Evaluation on Bernini-Bench.}
For fair comparison, the outputs of Kling O3 and Wan2.7 are downsampled to 480p at 16 fps, matching Bernini's generation setting. As shown in Table~\ref{tab:bernini_ben_res}, Bernini achieves the best overall performance on Bernini-V2V, raising the overall score from 3.30 to 3.49 compared with Wan2.7. Relative to Kling O3, it consistently outperforms across all evaluation dimensions. Relative to Wan2.7, Bernini is comparable in instruction following and generation quality, but shows a markedly stronger ability to preserve video consistency. On Bernini-RV2V, Bernini again achieves the best video consistency and remains competitive on the other metrics. These results show that Bernini is able to preserve consistency in non-edited regions to the greatest extent possible while correctly executing the instruction, which is often overlooked by existing editing models.

\begin{figure}[!t]
    \centering
    \includegraphics[width=0.9\linewidth]{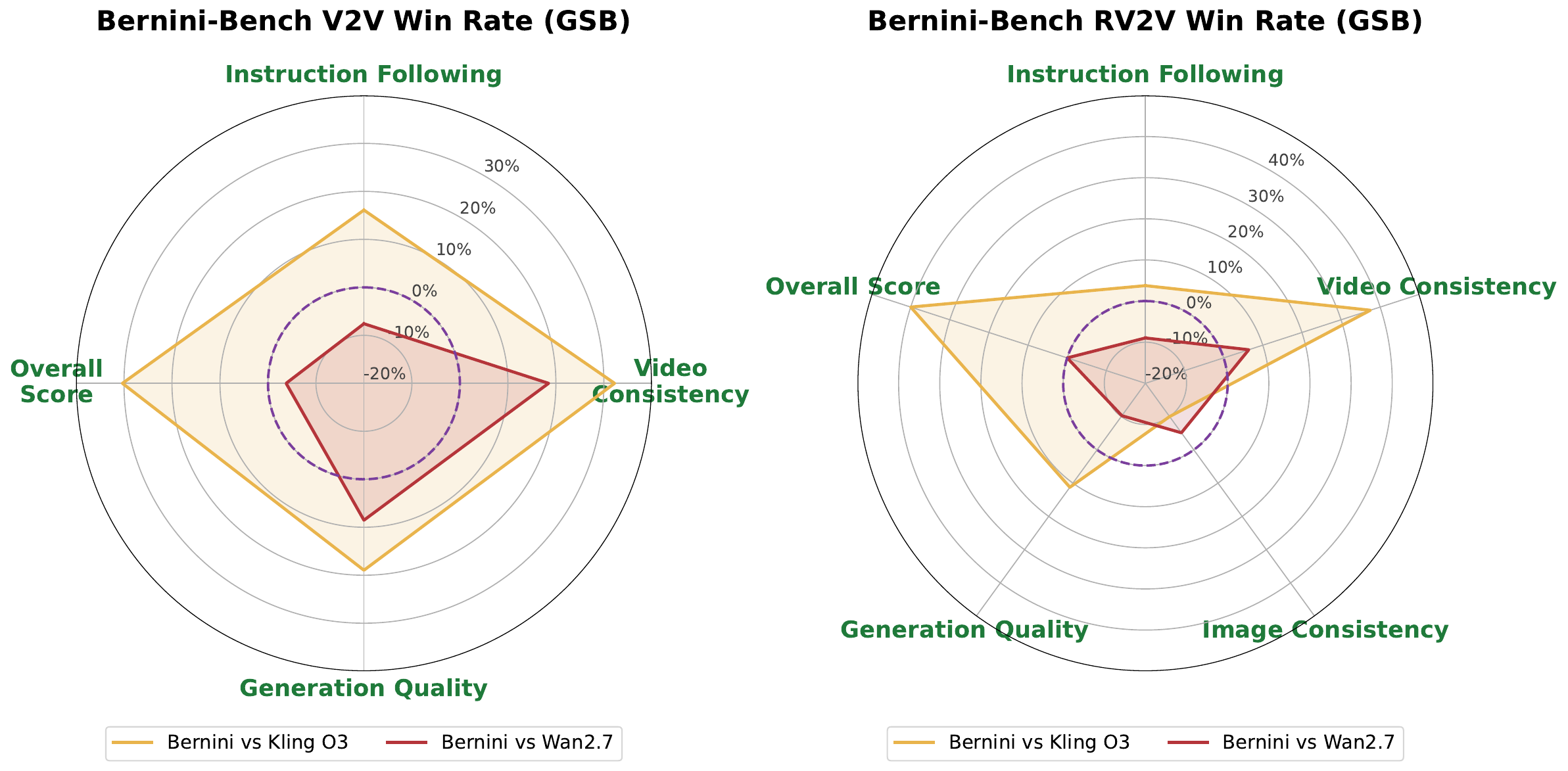}
    \caption{\textbf{GSB win rates on Bernini-Bench.} Relative win rates ((Good - Bad) / Total) of Bernini against Kling O3 and Wan2.7 on V2V (left) and RV2V (right).}
    \label{fig:bernini_bench_gsb_v2}
\end{figure}

\textbf{Human Side-By-Side Evaluation on Bernini-Bench.}
As shown in Fig.~\ref{fig:bernini_bench_gsb_v2}, we present the results of side-by-side (SBS) human evaluation conducted on the Bernini-Bench. Compared with the scores from MLLM-based automatic evaluation, human evaluation can more accurately reflect the actual performance of different models. Bernini outperforms Kling O3 across most evaluation dimensions, and achieves competitive performance on par with Wan2.7. In particular, Bernini exhibits a significant advantage in terms of Video Consistency.

\begin{table}[!t]
    \centering
    \caption{Quantitative Comparison on OpenVE-Bench with Gemini 2.5 Pro.}
    \label{tab:openve_gemini}
    \small
    \setlength{\tabcolsep}{3.5pt}
    \begin{tabular}{l | c c c c c c c c c} 
        \toprule
        Method & Overall $\uparrow$ & \makecell{Global \\ Style} & \makecell{Background \\ Change} & \makecell{Local \\ Change} & \makecell{Local \\ Remove} & \makecell{Local \\ Add} & \makecell{Subtitle \\ Edit} & \makecell{Creative \\ Edit} & \makecell{Camera \\ Edit}  \\
        \midrule
        VACE-14B~\cite{jiang2025vace}  & 1.57 & 1.49 & 1.55 & 2.07 & 1.46 & 1.26 & 1.48 & 1.47 & 1.62 \\
        OmniVideo~\cite{omnivideo} & 1.31 & 1.11 & 1.18 & 1.14 & 1.14 & 1.36 & 1.00 & 2.26 & 1.00 \\
        InsViE~\cite{wu2025insvie} & 1.53 & 2.20 & 1.06 & 1.48 & 1.36 & 1.17 & 2.18 & 2.02 & 1.09 \\
        Ditto~\cite{ditto} & 1.98 & 4.01 & 1.68 & 2.03 & 1.53 & 1.41 & 2.81 & 1.23 & 1.32  \\
        ICVE~\cite{icve} & 2.07 & 2.22 & 1.62 & 2.57 & 2.51 & 1.97 & 2.09 & 2.41 & 1.11 \\
        Lucy-Edit~\cite{lucyedit} & 2.15 & 2.27 & 1.57 & 3.20 & 1.75 & 2.30 & 1.61 & 2.86 & 1.61 \\
        OpenVE-Edit~\cite{he2025openve} & 2.49 & 3.16 & 2.36 & 2.98 & 1.85 & 2.15 & \underline{2.91} & 2.31 & 2.02 \\
        VINO~\cite{vino2026}  & \underline{3.18} & \textbf{4.34} & \underline{2.54} & \underline{3.73} & \underline{3.22} & \underline{2.77} & 2.61 & \underline{3.29} & \underline{2.81} \\
        \midrule
        Bernini & \textbf{3.96} & \underline{4.29} & \textbf{3.58} & \textbf{4.72} & \textbf{4.10} & \textbf{3.35} & \textbf{3.55} & \textbf{3.84} & \textbf{4.18} \\
        \bottomrule
    \end{tabular}
\end{table}

\begin{table}[!t]
    \centering
    \caption{Comparison of video editing methods on EditVerse.}
    \label{tab:video_edit_benchmark}
    \small
    \begin{tabular}{l|cccccc}
    \toprule
    \multirow{2}*{Method}
    & VLM evaluation
    & Video Quality
    & \multicolumn{2}{c}{Text Alignment}
    & \multicolumn{2}{c}{Temporal Consist.} \\
    \cmidrule(lr){2-2} \cmidrule(lr){3-3} \cmidrule(lr){4-5} \cmidrule(lr){6-7}
    & Editing Quality $\uparrow$
    & Pick Score $\uparrow$
    & Frame $\uparrow$
    & Video $\uparrow$
    & CLIP $\uparrow$
    & DINO $\uparrow$ \\
    \midrule
    TokenFlow~\cite{geyer2024tokenflow}   & 5.26 & 19.73 & 25.57 & 22.70 & 98.36 & 98.09 \\
    STDF~\cite{yatim2024stdf}             & 4.41 & 19.45 & 25.24 & 22.26 & 96.04 & 95.22 \\
    Se\~norita-2M~\cite{zi2025senorita}   & 6.97 & 19.71 & 26.34 & 23.24 & 98.05 & 97.99 \\
    InsV2V~\cite{cheng2023insv2v}         & 5.21 & 19.39 & 24.99 & 22.54 & 97.15 & 96.57 \\
    Lucy-Edit~\cite{lucyedit}             & 5.89 & 19.67 & 26.00 & 23.11 & 98.49 & \underline{98.38} \\
    EditVerse~\cite{ju2025editverse}      & \underline{7.65} & \underline{20.07} & \underline{26.73} & \underline{23.93} & \textbf{98.56} & \textbf{98.42} \\
    \midrule
    Bernini                     & \textbf{8.02} & \textbf{20.26} & \textbf{27.37} & \textbf{24.62} & \underline{98.55} & 98.37 \\
    \bottomrule
    \end{tabular}
\end{table}

\label{app:five_bench}
\begin{table}[!t]
\centering
\caption{Quantitative Comparison on FiVE Benchmark~\cite{li2025five}.}
\label{tab:five_bench}
\resizebox{\linewidth}{!}{%
    \begin{tabular}{l|c|ccc|cc|c|ccccc}
    \toprule
    \multirow{2}{*}{Method} & Structure & \multicolumn{3}{c|}{Background Preservation} & \multicolumn{2}{c|}{Text Alignment} & Motion & \multicolumn{5}{c}{FiVE} \\
    \cmidrule(lr){2-2}\cmidrule(lr){3-5}\cmidrule(lr){6-7}\cmidrule(lr){8-8}\cmidrule(lr){9-13}
     & Dist.$\times 10^3\downarrow$ & PSNR$\uparrow$ & LPIPS$\times 10^3\downarrow$ & SSIM$\times 10^2\uparrow$ & CLIPS.$\uparrow$ & CLIPS.$_\text{edit}\uparrow$ & Fid S.$\times 10^2\uparrow$ & YN$\uparrow$ & MC$\uparrow$ & $\cup\uparrow$ & $\cap\uparrow$ & Acc$\uparrow$ \\
    \midrule
    Source Videos & 0.00 & $\infty$ & 0.00 & 100.00 & 24.59 & 19.87 & 93.76 & -- & -- & -- & -- & -- \\
    \midrule
    TokenFlow~\cite{geyer2024tokenflow}   & 35.62 & 19.06 & 263.61 & 72.51 & 26.46 & 21.15 & 89.00 & 19.36 & 35.51 & 36.68 & 18.18 & 27.43 \\
    DMT~\cite{yatim2024stdf}               & 85.95 & 14.71 & 404.60 & 51.64 & 26.66 & \underline{21.44} & 82.30 & 34.78 & 62.06 & 62.98 & 33.86 & 48.42 \\
    VidToMe~\cite{li2024vidtome}       & 22.37 & 21.15 & 263.91 & 70.69 & 26.84 & 21.05 & \textbf{90.06} & 20.03 & 33.50 & 36.20 & 17.34 & 26.77 \\
    AnyV2V~\cite{ku2024anyv2v}         & 71.36 & 15.90 & 348.59 & 50.77 & 24.89 & 19.72 & 60.36 & 30.62 & 45.42 & 48.96 & 27.09 & 38.02 \\
    VideoGrain~\cite{yang2025videograin} & \textbf{12.40} & \textbf{27.05} & \underline{185.21} & 79.13 & 25.69 & 20.31 & 88.57 & 30.50 & 43.97 & 44.30 & 30.17 & 37.23 \\
    Pyramid-Edit~\cite{li2025five}  & 28.65 & 20.84 & 276.59 & 71.72 & 26.82 & 20.20 & 80.59 & 33.67 & 54.01 & 56.36 & 31.31 & 43.84 \\
    Wan-Edit~\cite{li2025five}      & \underline{12.53} & 25.57 & \textbf{94.61} & \underline{82.55} & 26.39 & 21.23 & \underline{89.43} & 41.41 & 52.53 & 55.72 & 38.22 & 46.97 \\
    Omni~\cite{yang2026contextunrolling} & 34.94 & 22.95 & 217.55 & 73.78 & \underline{26.92} & 21.19 & 84.22 & \underline{62.83} & \underline{81.81} & \underline{84.33} & \underline{60.23} & \underline{72.41} \\
    \midrule
    Bernini & 13.54 & \underline{26.35} & 207.89 & \textbf{84.38} & \textbf{27.75} & \textbf{22.74} & 86.27 & \textbf{71.33} & \textbf{84.98} & \textbf{87.67} & \textbf{68.65} & \textbf{78.16} \\
    \bottomrule
    \end{tabular}%
}
\end{table}

\textbf{Evaluation on Public Benchmarks.}
We evaluate Bernini on three public video editing benchmarks: OpenVE~\cite{he2025openve}, EditVerse~\cite{ju2025editverse}, and FiVE~\cite{li2025five}. As shown in Tables~\ref{tab:openve_gemini}, \ref{tab:video_edit_benchmark}, and~\ref{tab:five_bench}, Bernini consistently delivers strong performance across diverse evaluation settings. On OpenVE, it outperforms the strongest baseline VINO by a large margin in overall score (4.04 vs.\ 3.18). On EditVerse, Bernini achieves the best editing quality, pick score, and text alignment, especially setting a new high score of 8.02 on editing quality. On FiVE, Bernini attains state-of-the-art or near-state-of-the-art performance across structure, background, alignment, and editing accuracy metrics, with particularly strong results on the VQA-based FiVE-Acc metrics, indicating more faithful realization of the target edits.

\begin{figure}[!htbp]
    \centering
    \includegraphics[width=1.0\linewidth]{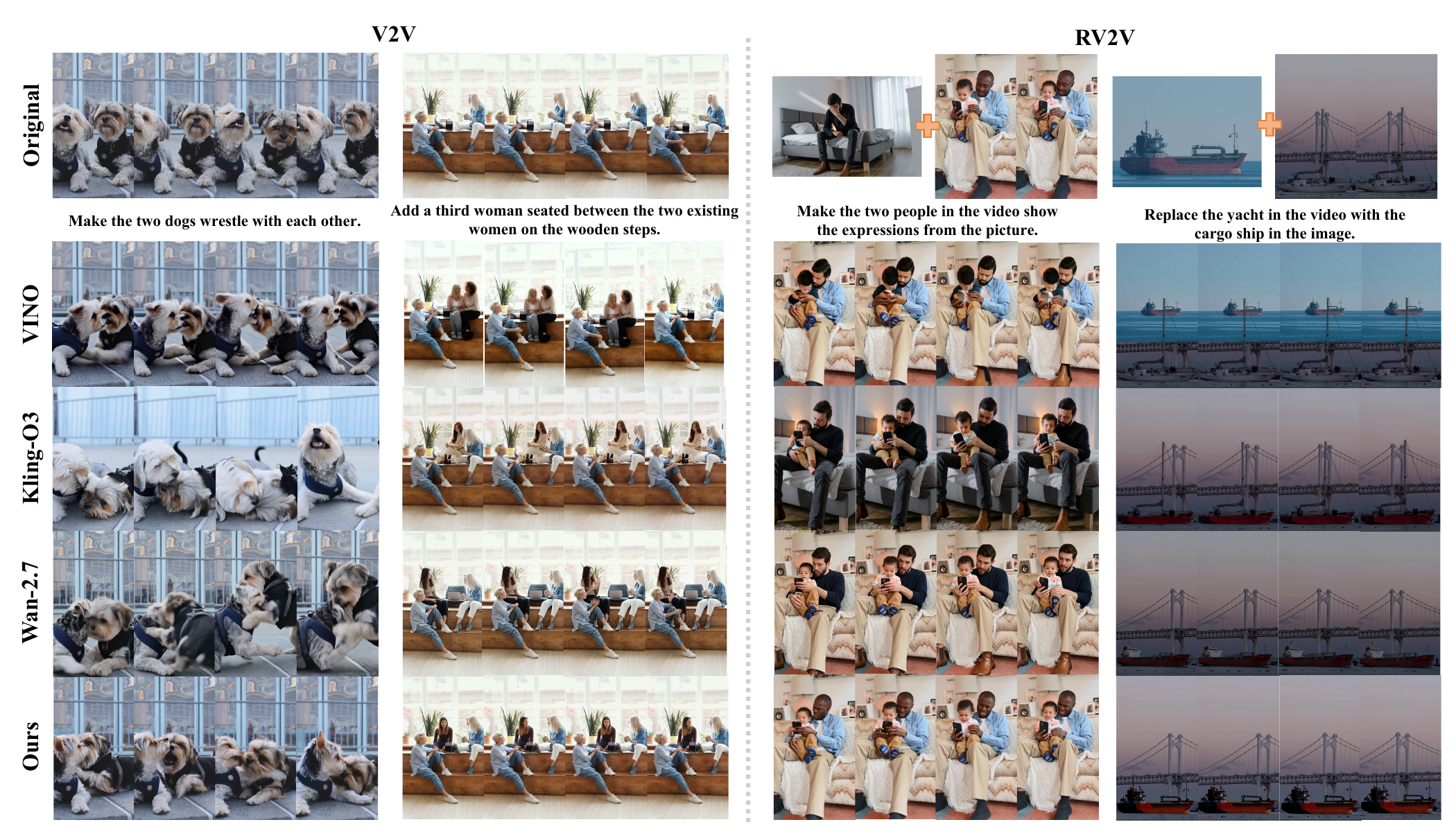}
    \caption{Qualitative comparison of Bernini with SoTA methods on V2V and RV2V tasks.}
    \label{fig:qualitative}
\end{figure}
\textbf{Qualitative Comparison.} Figure~\ref{fig:qualitative} presents a qualitative comparison between Bernini and existing state-of-the-art video editing models. In the V2V case of modifying the puppy’s motion, VINO produces a puppy that is inconsistent with the original video, Kling O3 generates a background that does not match the original video, and Wan2.7 produces obvious distortion in the puppy on the left. Only Bernini preserves both the puppy and the background consistently with the original video while naturally modifying the puppy’s motion. In the case of adding a person, Kling O3 successfully adds a person, but the added character sits stiffly. Both Wan2.7 and Bernini not only add a new person but also enable natural interaction between the new character and the two original girls; however, Wan2.7 incorrectly introduces an extra blue seat cushion in the middle. In the first RV2V case, only Bernini successfully preserves the identity consistency of the person in the video while modifying the facial expression. In the second RV2V case, only Bernini generates a boat that is consistent with the one in the reference image. More results are shown in Appendix~\ref{app:more_qualitative}.

\subsection{Reasoning-augmented Video Editing}
\label{sec:editing_with_reasoning}
\begin{table}[!t]
    \centering
    \small
    \caption{Comparison of reasoning variants on the Bernini-V2V benchmark. PE means a Prompt Enhancer that maps diverse user prompts onto a distribution
that is consistent with the model’s training data.}
    \label{tab:video_edit_benchmark_cot}
    \setlength{\tabcolsep}{4pt}
    \begin{tabular}{l|cccc}
        \toprule
        Method & OS & IF & VC & GQ \\
        \midrule
        Ours (baseline) & 3.12 & 3.36 & 3.18 & 3.37 \\
        \quad + PE (Qwen2.5-VL-7B~\cite{qwen2.5vl}) & 3.20 & 3.43 & 3.21 & 3.39 \\
        \midrule
        \addlinespace[2pt]
        \quad + Self-text & 3.33 & 3.55 & 3.31 & 3.44 \\
        \quad + PE (GPT-5.4) & \underline{3.49} & \textbf{3.66} & \underline{3.51} & \textbf{3.49} \\
        \quad + PE (GPT-5.4) + Self-vision-text & \textbf{3.52} & \underline{3.65} & \textbf{3.54} & \textbf{3.49} \\
        \bottomrule
    \end{tabular}

\end{table}

Table~\ref{tab:video_edit_benchmark_cot} presents the results on the Bernini-V2V benchmark.
Overall, enriching the reasoning context consistently improves performance.
First, we use Qwen2.5-VL-7B, the initialization model of our MLLM planner, as a Prompt Enhancer to refine the input prompts.
Although this brings slight improvements over the baseline, it performs worse than our self-text reasoning approach, indicating that our model develops stronger textual reasoning capabilities beyond its initialization model.
To further explore the upper bound of textual reasoning, we employ a stronger Prompt Enhancer GPT-5.4, which achieves consistent gains across most metrics.
Building on this, incorporating vision-text reasoning achieves the best overall performance, showing that multimodal reasoning provides complementary benefits beyond text-only reasoning.
Detailed qualitative comparisons are provided in Appendix~\ref{app:vis_cot}.

\subsection{Video Generation}
\label{sec:video_generation}

\begin{table}[!t]
  \caption{Quantitative comparison on VBench~\cite{vbench}.}
  \label{tab:vbench}
  \centering
  \small
  \setlength{\tabcolsep}{3.5pt}
  \begin{tabular}{lccccccc}
    \toprule
    Method & Total $\uparrow$ & \makecell{Quality\\score} & \makecell{Semantic\\score} & \makecell{Aesthetic\\quality} & \makecell{Dynamic\\degree} & \makecell{Object\\class} & \makecell{Overall\\consist.} \\
    \midrule
    \multicolumn{8}{l}{\textit{Closed-source systems}} \\
    Sora~\cite{sora}             & 84.28 & 85.51 & 79.35 & 63.46 & \textbf{79.91} & 93.93 & 26.26 \\
    Veo3~\cite{veo3}             & \textbf{85.06} & \textbf{85.70} & \underline{82.49} & 63.81 & 72.43 & 93.89 & \textbf{27.88} \\
    Kling~1.6~\cite{kling2024}       & 83.40 & 85.00 & 76.99 & 64.81 & 62.22 & 93.34 & 26.04 \\
    Jimeng~\cite{jimeng}         & 81.97 & 83.29 & 76.69 & \textbf{68.80} & 38.43 & 89.62 & 27.10 \\
    Gen-3~\cite{gen3}            & 82.32 & 84.11 & 75.17 & 63.34 & 60.14 & 87.81 & 26.69 \\
    \midrule
    \multicolumn{8}{l}{\textit{Open-source systems}} \\
    StepVideo~\cite{stepvideo}       & 81.83 & 84.46 & 71.28 & 61.23 & 53.06 & 80.56 & 27.12 \\
    CogVideoX-5B~\cite{cogvideox}    & 81.91 & 83.05 & 77.33 & 61.88 & 69.51 & 85.07 & 27.65 \\
    Wan2.1-14B~\cite{wan}         & 83.69 & \underline{85.59} & 76.11 & 66.07 & 65.46 & 86.28 & 25.91 \\
    HunyuanVideo~\cite{hunyuanvideo} & 83.24 & 85.09 & 75.82 & 60.36 & 70.83 & 86.10 & 26.44 \\
    VINO~\cite{vino2026}            & 83.17 & 83.69 & 81.08 & \underline{68.11} & 55.56 & 91.17 & 27.00 \\
    \midrule
    Wan2.2-A14B & \underline{84.79} & 85.33 & \textbf{82.61} & 67.06 & 69.72 & \textbf{96.00} & 27.36 \\
    Bernini & 84.46 & 85.20 & 81.49 & 65.32 & \underline{79.72} & \underline{95.08} & \underline{27.73} \\
    \bottomrule
  \end{tabular}
\end{table}

\textbf{Text-to-Video Generation.}
We evaluate Bernini on VBench~\cite{vbench} to assess its text-to-video generation capability. As Bernini is built on top of Wan2.2-A14B and extends it to a unified framework that additionally supports video editing and reference-to-video generation, we compare against Wan2.2-A14B to examine how the text-to-video capability is retained after this extension. As shown in Table~\ref{tab:vbench}, Bernini reaches a Total score of 84.64, essentially matching Wan2.2-A14B (84.79).
These results indicate that the unified design of Bernini broadens the model's capability across editing and reference-based generation tasks while retaining the base text-to-video quality.

\begin{table}[!t]
  \caption{OpenS2V open-domain results on reference-to-video generation.
  Higher is better for all metrics.}
  \label{tab:opens2v_full}
  \centering
  \small
  \setlength{\tabcolsep}{3.5pt}
  \begin{tabular}{lcccccccc}
   \toprule
    Method & Total & Aesth. & \makecell{Motion\\Smooth.} & \makecell{Motion\\Ampl.} & FaceSim & GmeScore & NexusScore & NaturalScore \\
    \midrule
    \multicolumn{9}{l}{\textit{Closed-source systems}} \\
    Pika~2.1~\cite{pika2024}              & 51.88 & 46.88 & 87.06          & 24.71          & 30.38          & 69.19          & 45.40 & 63.32 \\
    Vidu~2.0~\cite{vidu2024}              & 51.95 & 41.48 & 90.45          & 13.52          & 35.11          & 67.57          & 43.37             & 65.88 \\
    Kling~1.6~\cite{kling2024}            & 56.23 & 44.59 & 86.93          & \textbf{41.60} & 40.10          & 66.20          & 45.89    & \textbf{74.59} \\
    Kling~O3~\cite{kling2024} & 59.19 & \textbf{48.05} & 92.94 & 24.47 & \underline{57.20} & 66.44 & 45.53 & 70.51\\
    \midrule
    \multicolumn{9}{l}{\textit{Open-source systems}} \\
    SkyReels-A2~\cite{fei2025skyreelsa2}  & 52.25 & 39.41 & 87.93          & 25.60          & 45.95          & 64.54          & 43.75             & 60.32 \\
    MAGREF~\cite{deng2025magref}          & 52.51 & 45.02 & 93.17          & 21.81          & 30.83          & \underline{70.47}          & 43.04             & 66.90 \\
    Phantom-14B~\cite{liu2025phantom}     & 56.77 & 46.39 & \underline{96.31} & \underline{33.42}          & 51.46          & \textbf{70.65} & 37.43          & 69.35 \\
    VACE-14B~\cite{jiang2025vace}         & 57.55 & \underline{47.21} & 94.97 & 15.02 & 55.09 & 67.27       & 44.08             & 67.04 \\
    VINO~\cite{vino2026}                  & 57.85 & 45.92 & 94.73 & 12.30       & 52.00          & 69.69          & 42.67             & 71.99 \\
    Saber~\cite{zhou2025scaling} & 57.91 & 42.42 & 96.12 & 21.12 & 49.89 & 67.50 & 47.22 & 72.55 \\
    RefAlign-14B~\cite{wang2026refalign} & \underline{60.42} & 46.84 & \textbf{97.61} & 22.48 & 55.23 & 68.32 & \underline{48.52} & 73.63 \\
    \midrule
    Bernini & \textbf{63.83}   & 44.54    & 94.39           & 14.29           & \textbf{77.34}            & 64.15             & \textbf{49.21}             & \underline{73.94} \\
    \bottomrule
  \end{tabular}
\end{table}
\textbf{Reference-to-Video Generation.}
We evaluate our method on OpenS2V-Eval~\cite{yuan2025opens2v}, a benchmark for multi-reference reference-to-video generation spanning humans, objects, and face-identity consistency. Following its evaluation protocol, we report the overall Total score together with all sub-metrics defined therein.
As summarized in Table~\ref{tab:opens2v_full}, our method achieves the highest Total score of \textbf{62.94}, surpassing all closed-source and open-source competitors, including the strongest prior results from Kling~O3 (59.19) and RefAlign-14B (60.42). Most notably, our approach attains a FaceSim score of \textbf{78.20}, exceeding the next-best baseline Kling~O3 (57.20) by over 20 absolute points. This pronounced margin demonstrates substantially stronger face-identity preservation, which has long been a key challenge in multi-reference subject-driven video generation.

\subsection{Ablation Studies}
\label{sec:ablation}

\begin{figure*}[!t]
    \centering
    \includegraphics[width=0.98\linewidth]{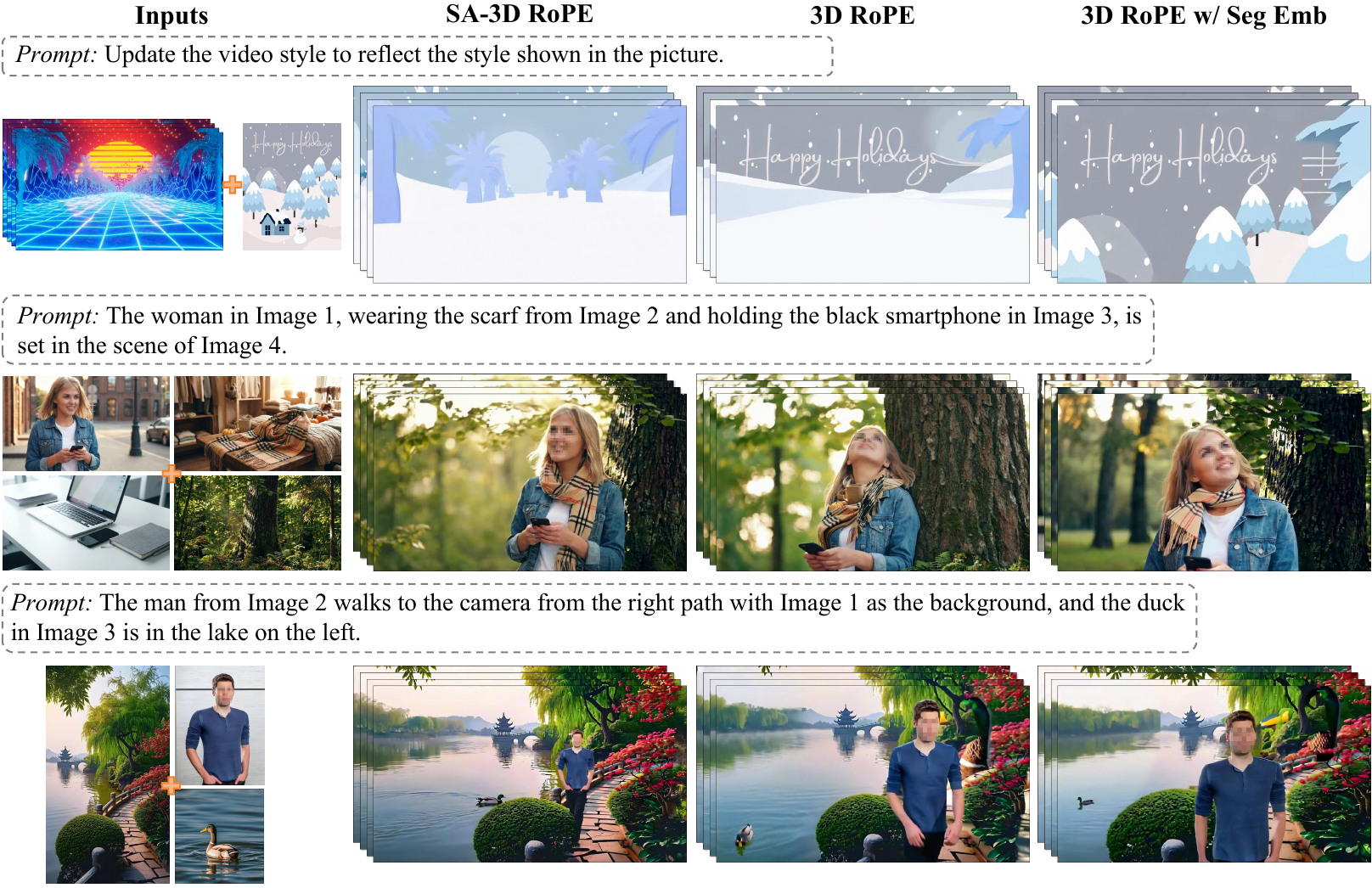}
    \caption{Ablation study of SA-3D RoPE, standard 3D RoPE and 3D RoPE with segment embedding on reference-guided video editing and reference-to-video tasks. Although incorporating segment embedding brings improvements in reference consistency (e.g., the scarf in the 2nd row), both baseline variants still suffer from noticeable reference leakage artifacts (e.g., the background in the 1st row and the duck head in the 3rd row).}
    \label{fig:ablation_rope}
\end{figure*}

\textbf{Effect of SA-3D RoPE.} 
We further examine the contribution of segment-aware position encoding by comparing SA-3D RoPE against two baselines in reference-based video editing: the standard 3D RoPE, and the standard 3D RoPE with learnable segment embeddings. In this unified sequence, target tokens, source video tokens, and reference image tokens coexist. For the segment embedding baseline, segment embeddings are added to the hidden states at each DiT layer, utilizing the same segment IDs as SA-3D RoPE. As shown in Fig.~\ref{fig:ablation_rope}, while the explicit segment embeddings improve reference consistency over the vanilla 3D RoPE, both baselines fail to cleanly isolate features. Consequently, they both suffer from content confusion, causing appearance details from the reference image to leak into unintended regions of the target. %
This confirms that additive segment embeddings are insufficient when multiple visual segments share the same (t, h, w) coordinates, so the renderer cannot reliably distinguish their roles. SA-3D RoPE introduces a segment-index-conditioned phase modulation that decouples segment identity from spatiotemporal position, allowing attention to attend to the correct segment while preserving the original spatiotemporal modeling properties of 3D RoPE.

\begin{figure*}[htbp]
    \centering
    \includegraphics[width=0.98\linewidth]{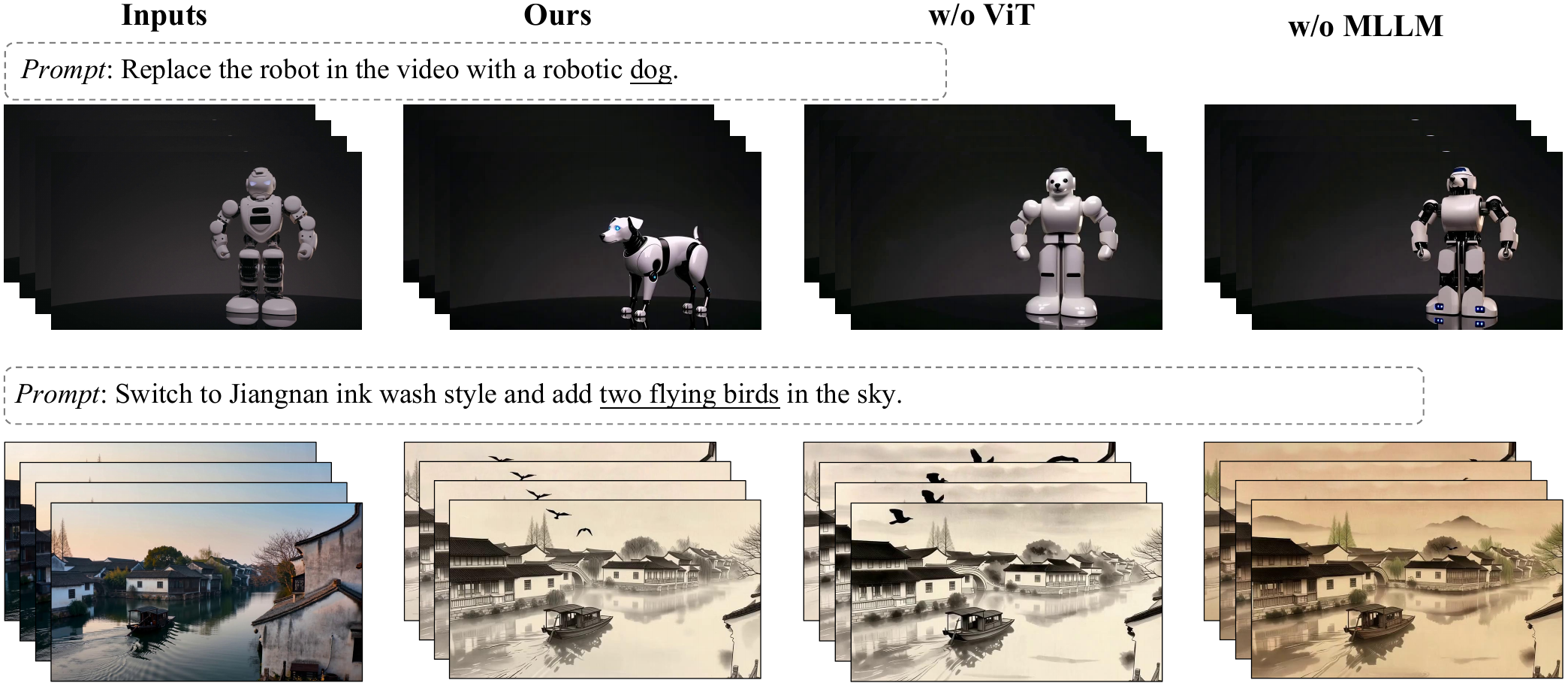}
    \caption{Ablation study on the ViT semantic interface and the MLLM planner.}
    \label{fig:ablate_clip}
\end{figure*}
\begin{figure*}[t]
    \centering
    \includegraphics[width=0.92\linewidth]{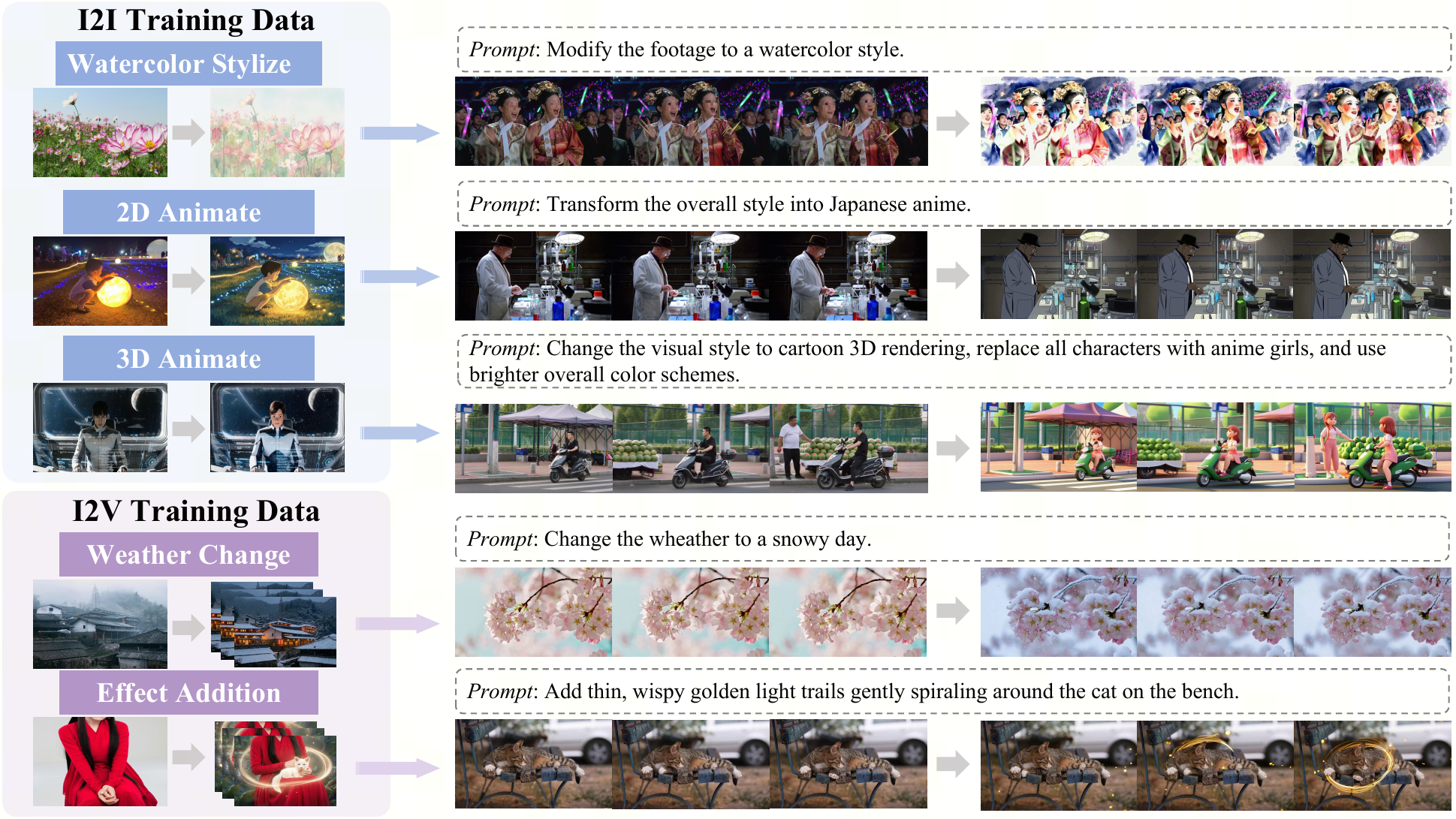}
    \caption{Video editing generalization through diverse I2I and I2V training data.}
    \label{fig:i2ii2v_generalize}
\end{figure*}

\textbf{Effect of ViT Semantic Interface and MLLM Planner.}
As shown in Fig.~\ref{fig:ablate_clip}, both the ViT embedding decoder and the MLLM planner are crucial for high-quality video editing. Our full model accurately performs object replacement and style transfer while preserving scene consistency. Removing the ViT semantic interface leads to weaker instruction following, such as failing to replace the robot with a robotic dog or omitting the flying birds in the Jiangnan ink wash style editing. Removing both ViT and MLLM further degrades the results, demonstrating their complementary roles in precise and faithful editing.

\subsection{Generalizations}
\label{sec:generalization}
\begin{figure*}[h]
    \centering
    \includegraphics[width=0.87\linewidth]{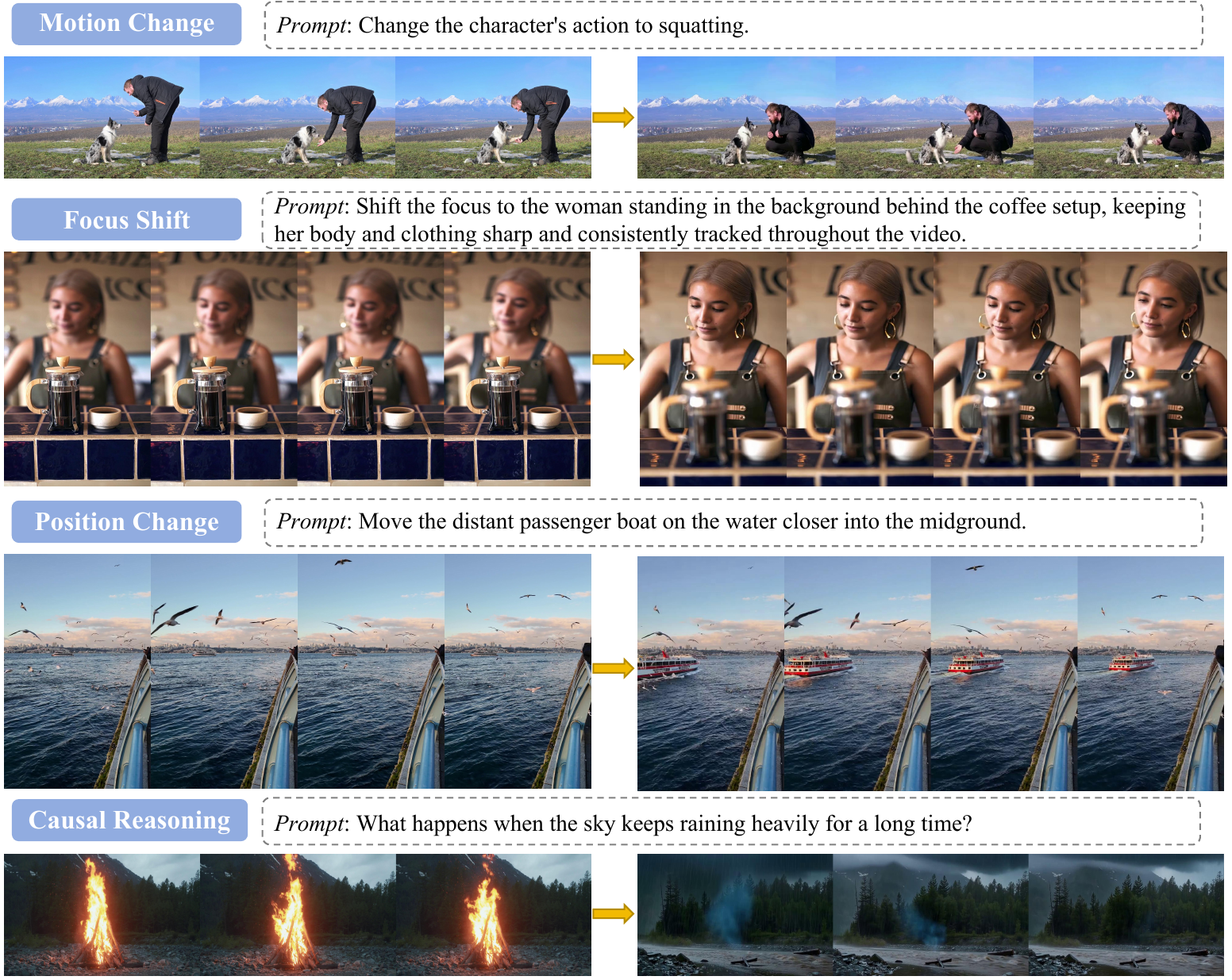}
    \caption{Generalization to diverse video editing instructions.}
    \label{fig:generalize}
\end{figure*}
As shown in Figs.~\ref{fig:i2ii2v_generalize} and~\ref{fig:generalize}, Bernini demonstrates strong generalization to diverse video editing instructions. 
Benefiting from heterogeneous I2I and I2V training data, Bernini successfully transfers the learned instruction-following capability to video editing scenarios, including watercolor stylization, 2D/3D animation, weather changes, and effect additions, as illustrated in Fig.~\ref{fig:i2ii2v_generalize}.

Furthermore, Bernini can handle editing instructions that are not explicitly present in the training data, such as motion changes, focus shifts, position changes, and causal reasoning, as shown in Fig.~\ref{fig:generalize}. 
Moreover, it supports reasoning-based editing: given the prompt about prolonged heavy rain, Bernini correctly infers that the fire should be extinguished, despite the absence of explicit causal supervision in the training data. These results suggest that Bernini does not merely memorize or fit standard training transformations, but instead learns a transferable and compositional instruction-following ability for video editing.

\section{Related Work}
\textbf{Joint Multimodal Backbones.}
One line of work merges understanding and generation into a single backbone that processes text and visual tokens together over a unified sequence. Emu3~\cite{emu3} embodies the simplest form of this idea, tokenizing text, images, and videos into a shared discrete vocabulary and training a single transformer from scratch with pure next-token prediction. Janus~\cite{janus} retains the unified autoregressive backbone but decouples the visual encoders for understanding and generation into two separate pathways, using a SigLIP-style encoder for perception and a VQ tokenizer for synthesis. Other works hybridize the modeling objective across modalities: Show-o~\cite{showo} couples autoregressive text modeling with discrete diffusion over image tokens within a single transformer, and HunyuanImage 3.0~\cite{hunyuanimage3} extends this hybrid to a Mixture-of-Experts decoder that performs next-token prediction for text alongside diffusion-based prediction for image tokens. BAGEL~\cite{bagel} adopts a Mixture-of-Transformer-Experts architecture in which separate understanding and generation experts interact through shared self-attention, paired with dual visual encoders that capture pixel- and semantic-level features. Lumina-DiMOO~\cite{lumina-dimoo} replaces autoregressive prediction altogether with fully discrete masked diffusion as a single training objective over both modalities under a shared vocabulary.

\textbf{MLLMs as Conditioners for Visual Generation.}
A second line of work keeps the MLLM and the diffusion model as separate components and lets the MLLM provide conditioning signals, with works differing primarily in what representation carries the signal. The narrowest interfaces pass the MLLM's output text tokens or a small group of learnable query tokens into the diffusion model through cross-attention: MetaQuery~\cite{metaquery} and Bifrost-1~\cite{bifrost} fall into this category, with Bifrost-1 specifically using patch-level CLIP latents as the bridge. Wider interfaces use the MLLM's hidden states directly: SEED-X~\cite{seedx}, DreamLLM~\cite{dreamllm}, and Emu~\cite{emu} drive an external image decoder from these hidden states, while LaVi-Bridge~\cite{lavibridge} more broadly focuses on connecting frozen language and vision generators. The same hidden-state interface has been extended to video by UniVideo~\cite{univideo} and VINO~\cite{vino2026}, which couple an MLLM with a video diffusion backbone and feed MLLM hidden states (optionally augmented with learnable query tokens or interleaved multimodal context, alongside VAE latents of references) into the generator, unifying reference-to-video~\cite{jiang2025vace,liu2025phantom,fei2025skyreelsa2,deng2025magref,wang2026refalign} and instruction-based editing~\cite{wu2025insvie,lucyedit,ditto,he2025openve} under a single framework. Our work, Bernini, follows this decoupled paradigm but anchors the interface to the MLLM's own ViT embedding space rather than its output hidden states, so that pretrained visual semantics can be transferred to the diffusion renderer at their native representation.

\section{Conclusion and Limitations}
We presented Bernini, a unified framework for video generation and editing that decouples semantic planning from pixel rendering: an MLLM planner predicts the target in its native ViT embedding space, and a DiT renderer synthesizes pixels conditioned on this plan, augmented by text and source VAE features. This interface lets the two components be trained largely independently while preserving their pretrained strengths. With SA-3D RoPE for multi-segment disambiguation and a latent chain-of-thought planner, Bernini achieves state-of-the-art results across video editing and reference-to-video benchmarks, and generalizes to challenging instructions beyond standard training cases.

Bernini remains much limited by our adopted foundation models for both MLLM planner and DiT renderer. In complex editing scenarios, it still depends on a strong LLM rewriter to provide sufficiently detailed and structured instructions, indicating that its native reasoning ability is not yet fully sufficient for challenging edits. In addition, while Bernini achieves state-of-the-art consistency in reference-to-video generation, its visual quality still falls short of stronger closed-source systems such as Wan2.7. More powerful foundation model instantiations could be helpful to further improve results.
\section{Contributions and Acknowledgements}
Authors are organized by contribution role. All algorithm authors contributed equally to this work and are listed in alphabetical order by first name. $^\dag$ indicates the Project Lead.

\textbf{Algorithm:}
Chenchen Liu, Junyi Chen, Lei Li, Lu Chi$^\dag$, Mingzhen Sun, Zhuoying Li

\textbf{Infrastructure:}
Yi Fu, Ruoyu Guo, Yiheng Wu, Ge Bai

\textbf{Team Leader:}
Zehuan Yuan

We would like to thank Ruibiao Lu, Mingyang Zou and Zhen Ye for their support throughout this project.

\clearpage

\bibliographystyle{plainnat}
\bibliography{egbib}

\clearpage
\beginappendix

\section{MLLM Prompts for Bernini-Bench Evaluation}
\label{app:eval_prompts}

\lstdefinestyle{promptstyle}{
    basicstyle=\ttfamily\small,
    backgroundcolor=\color{gray!10},
    frame=single,
    breaklines=true,          %
    breakindent=0pt,          %
    breakatwhitespace=false,
    columns=fullflexible,
    keepspaces=true,
    showstringspaces=false,
    literate={“}{{``}}1 {”}{{''}}1 {‘}{{`}}1 {’}{{'}}1  %
}

\begin{lstlisting}[style=promptstyle, title={\textbf{Prompt 1.} Bernini-V2V evaluation prompt}]
You are a professional data rater specializing in evaluating instruction-driven video editing results. You will be given two videos (the original video before editing and the result video after editing) along with the corresponding editing instruction. Your task is to evaluate the editing quality on a 1-5 scale across four dimensions: Instruction Compliance, Video Consistency, Generation Quality, and Overall Performance.

# Dimension 1 Instruction Compliance

Evaluate whether the editing instruction has been faithfully and completely executed. Focus on: Was the correct target edited? Is the action type correct (add / remove / replace / style transfer / background change / camera change, etc.)? Are all specified attributes (object class, count, position, colour, size, style, etc.) satisfied?

Special rule - "Non-execution": If the target video shows NO meaningful change from the original (minor colour shifts on non-colour instructions, or trivial changes in non-target areas do NOT count as execution), treat it as instruction non-compliance and score 1.

1 - Non-execution or completely wrong: The edited video shows NO meaningful change from the original video (instruction completely ignored, or the video is an identical copy; minor colour shifts on non-colour instructions or trivial irrelevant fluctuations do NOT count as execution), OR the video is corrupted / the edit is entirely wrong (wrong target, wrong action type, entirely unrelated change, or the foreground/background is destroyed in a way that has nothing to do with the instruction).
2 - The edit partially attempts the instruction but fundamentally fails: only a few frames are affected, the wrong object/class is edited, the action is applied to the wrong region, or the instruction is executed in a clearly incorrect way (e.g., object added on the wrong side, wrong style applied).
3 - The core instruction is largely followed (correct target, correct action type), but with significant errors: key attributes are wrong (e.g., count, position, colour clearly deviate from the prompt), remnants of removed objects remain, unintended objects are also edited, or the effect is highly inconsistent across frames.
4 - The instruction is correctly and fully executed for the entire duration. Only minor attribute inaccuracies remain (e.g., slight colour or size mismatch, minor position offset). All and only the intended targets are affected.
5 - Perfect execution: every aspect of the instruction is faithfully reproduced - target, action, class, number, position, scale, pose, motion, style, and detail all exactly match the prompt throughout the entire video.

# Dimension 2 Video Consistency

Evaluate whether non-edited regions and editing-irrelevant attributes remain consistent before and after editing. This includes: background preservation, art style / colour tone consistency, subject identity consistency (same person, same object), spatial layout and geometry preservation, and motion continuity.

Important: Changes that are a NECESSARY and EXPECTED consequence of the editing instruction itself should NOT be penalised in this dimension (e.g., a camera shot change naturally alters the visible background; a style transfer naturally changes the colour palette). Only evaluate unintended or unnecessary deviations.

1 - The original scene is barely recognisable: background is completely replaced or destroyed, subject identity is lost, spatial layout is severely distorted, or art style is drastically and unnecessarily altered.
2 - The main subject is recognisable, but major unintended changes are present: background is significantly altered, subject's key features (shape, appearance, identity) are clearly different, or the overall colour tone / art style shifted noticeably without instruction.
3 - Overall structure and subject identity are maintained, but noticeable unintended deviations exist: moderate background changes, minor subject appearance drift (e.g., clothing colour shift, slight shape change), or mild art style inconsistency in some frames.
4 - Nearly all non-edited regions and attributes are well preserved. Only very minor, hard-to-spot deviations (e.g., subtle texture change in a small background area, very slight colour tone shift) that do not affect the overall viewing experience.
5 - Perfect preservation: all non-edited regions, subject identity, background, art style, spatial layout, and motion are completely unchanged. The edit is perfectly isolated to the instructed target.

# Dimension 3 Generation Quality

Evaluate the visual quality, temporal stability, and physical plausibility of the edited video. Focus on: AI artifacts (pasting feel, unnatural blending, distortion, deformation), temporal coherence (flickering, jittering, "boiling" textures, frame discontinuity), seamlessness of integration (edges, colour/resolution matching, lighting/shadow consistency), and physical realism (correct perspective, shadows, reflections, occlusion, natural motion).

1 - Severe quality issues: extreme flickering or "boiling" effects, heavy distortion/deformation of generated content, obvious pasting feel with clear seams, or physically impossible results (floating objects, completely wrong perspective/lighting). The video is essentially unwatchable.
2 - Significant and distracting quality problems: obvious AI artifacts, strong temporal inconsistency (style flickers on/off, jittering edges), clear resolution/colour mismatch between edited and unedited regions, or major physical implausibilities (missing/static shadows, wrong occlusion).
3 - Noticeable but tolerable issues: moderate AI feel (slightly unnatural blending), some flickering or texture instability during motion, minor edge artefacts or colour bleeding, or small physical inconsistencies (slightly off shadows or perspective). The video is watchable but the edit is clearly visible upon normal viewing.
4 - Good quality with only minor issues: very slight AI artifacts visible only upon close inspection, largely stable with only subtle flickering in complex motion areas, well-matched lighting and colour, and believable physical interactions. Casual viewers would not notice the edit.
5 - Flawless quality: perfectly stable and temporally coherent with zero flickering, completely seamless integration indistinguishable from the original footage, physically correct lighting/shadows/reflections/perspective throughout, and natural motion. The edit appears as if it were part of the original recording.

# Dimension 4 Overall Performance

Provide a holistic assessment: considering all the above dimensions together, how well does the edited video meet the user's likely expectation? This score reflects the overall subjective satisfaction - whether the result would be accepted by a real user as a successful edit.

1 - Completely fails to meet user expectations. The edit is unusable.
2 - The edit attempt is recognisable but the result is clearly unacceptable due to combined failures in compliance, consistency, or quality.
3 - A partially successful edit: the intent is understood and partially achieved, but noticeable issues in one or more dimensions reduce the result's usability.
4 - A good edit that would satisfy most users. Minor imperfections exist but do not significantly detract from the overall result.
5 - An excellent edit that fully meets or exceeds user expectations. The result is professional-grade and virtually indistinguishable from a manually crafted or real video.

# Scoring Constraints

- If the instruction is not executed at all (non-execution), Instruction Compliance should be scored 1, and Video Consistency and Generation Quality should be skipped (marked as "N/A"). Overall Performance should be 1.

# Response Format

Please output a valid JSON object exactly matching the following structure, without any extra markdown formatting or conversational text:
{{
  "Brief reasoning": "<A concise explanation covering all four dimensions, no more than 50 words.>",
  "Instruction Compliance": "<1-5>",
  "Video Consistency": "<1-5 or \"N/A\">",
  "Generation Quality": "<1-5 or \"N/A\">",
  "Overall Performance": "<1-5>"
}}

editing instruction is: {edit_prompt}.
Below are the videos before and after editing:
\end{lstlisting}

\begin{lstlisting}[style=promptstyle, title={\textbf{Prompt 2.} Bernini-RV2V evaluation prompt}]
You are a professional data rater specializing in evaluating reference-guided video editing results. You will be given an original video (before editing), a reference image (the visual exemplar that the edit should follow), the result video (after editing), and the corresponding editing instruction. Your task is to evaluate the editing quality on a 1-5 scale across five dimensions: Instruction Compliance, Video Consistency, Reference Image Consistency, Generation Quality, and Overall Performance.

# Dimension 1 Instruction Compliance

Evaluate whether the editing instruction has been faithfully and completely executed, INDEPENDENT of the reference image. Focus purely on whether the correct action was performed on the correct target: Was the right object/region edited? Is the action type correct (add / remove / replace / change background / change style, etc.)? Are positional and structural requirements satisfied?

Note: The degree to which the result matches the reference image's appearance is evaluated separately in Dimension 3. Here, only evaluate the structural and semantic correctness of the editing action itself.

Special rule - "Non-execution": If the target video shows NO meaningful change from the original (minor colour shifts on non-colour instructions, or trivial changes in non-target areas do NOT count as execution), treat it as instruction non-compliance and score 1.

1 - Non-execution or completely wrong: The edited video shows NO meaningful change from the original video (instruction completely ignored, or the video is an identical copy; minor colour shifts on non-colour instructions or trivial irrelevant fluctuations do NOT count as execution), OR the video is corrupted / the edit is entirely wrong (wrong target, wrong action type, entirely unrelated change, or the foreground/background is destroyed in a way that has nothing to do with the instruction).
2 - The edit partially attempts the instruction but fundamentally fails: only a few frames are affected, the wrong object/class is edited, the action is applied to the wrong region, or the instruction is executed in a clearly incorrect way (e.g., object added on the wrong side, wrong style applied).
3 - The core instruction is largely followed (correct target, correct action type), but with significant errors: key attributes are wrong (e.g., count, position, colour clearly deviate from the prompt), remnants of removed objects remain, unintended objects are also edited, or the effect is highly inconsistent across frames.
4 - The instruction is correctly and fully executed for the entire duration. Only minor attribute inaccuracies remain (e.g., slight colour or size mismatch, minor position offset). All and only the intended targets are affected.
5 - Perfect execution: every aspect of the instruction is faithfully reproduced - target, action, class, number, position, scale, pose, motion, style, and detail all exactly match the prompt throughout the entire video.

# Dimension 2 Video Consistency

Evaluate whether non-edited regions and editing-irrelevant attributes remain consistent before and after editing. This includes: background preservation, art style / colour tone consistency, subject identity consistency, spatial layout and geometry preservation, and motion continuity.

Important: Changes that are a NECESSARY and EXPECTED consequence of the editing instruction itself should NOT be penalised (e.g., replacing the background naturally changes the background; changing material naturally alters texture). Only evaluate unintended or unnecessary deviations.

1 - The original scene is barely recognisable: background is completely destroyed, subject identity is lost, spatial layout is severely distorted, or art style is drastically and unnecessarily altered.
2 - The main subject is recognisable, but major unintended changes are present: background significantly altered without reason, subject's key features clearly different, or overall colour tone shifted noticeably.
3 - Overall structure and subject identity are maintained, but noticeable unintended deviations exist: moderate background changes, minor subject appearance drift, or mild art style inconsistency.
4 - Nearly all non-edited regions and attributes are well preserved. Only very minor, hard-to-spot deviations that do not affect the overall viewing experience. Motion is smooth and continuous.
5 - Perfect preservation: all non-edited regions, subject identity, background, art style, spatial layout, and motion are completely unchanged. The edit is perfectly isolated.

# Dimension 3 Reference Image Consistency

Evaluate how well the edited result matches the reference image in terms of visual appearance. This is NOT a pixel-level comparison - it should be evaluated in the context of the editing instruction's intent. Focus on: Does the edited content capture the key visual characteristics of the reference image (shape, colour, texture, pattern, style, material, identity)? Is the resemblance faithful enough to recognise the reference as the source of inspiration?

Important: The reference image serves as a visual exemplar. The edit does not need to be a literal copy-paste - it should integrate the reference's visual characteristics naturally into the video scene while respecting the editing instruction's intent. A result that captures the essence (e.g., correct material/texture for a material change, correct object identity for a replacement, correct style for a style transfer) should score highly even if minor details differ.

1 - The edited content bears no resemblance to the reference image. The visual characteristics (shape, colour, texture, style, identity) are entirely different or absent.
2 - Very weak resemblance: only one or two superficial attributes vaguely match (e.g., similar general colour but wrong shape/texture/identity). The reference is not recognisable as the source.
3 - Moderate resemblance: the general category or style is correct, and some key visual features match, but significant differences remain in important attributes (e.g., correct object type but wrong colour/pattern, correct style direction but inconsistent execution).
4 - Strong resemblance: the edited content clearly reflects the reference image's key visual characteristics. Most attributes (shape, colour, texture, pattern, style, identity) are well captured. Only minor detail differences exist.
5 - Excellent match: the edited content faithfully reproduces the reference image's visual characteristics within the video context. All key attributes are accurately captured, and the integration feels natural and intentional.

# Dimension 4 Generation Quality

Evaluate the visual quality, temporal stability, and physical plausibility of the edited video. Focus on: AI artifacts (pasting feel, unnatural blending, distortion, deformation), temporal coherence (flickering, jittering, "boiling" textures, frame discontinuity), seamlessness of integration (edges, colour/resolution matching, lighting/shadow consistency), and physical realism (correct perspective, shadows, reflections, occlusion, natural motion).

1 - Severe quality issues: extreme flickering or "boiling" effects, heavy distortion/deformation, obvious pasting feel with clear seams, or physically impossible results. The video is essentially unwatchable.
2 - Significant and distracting quality problems: obvious AI artifacts, strong temporal inconsistency, clear resolution/colour mismatch, or major physical implausibilities.
3 - Noticeable but tolerable issues: moderate AI feel, some flickering or texture instability, minor edge artefacts, or small physical inconsistencies. The video is watchable but the edit is clearly visible.
4 - Good quality with only minor issues: very slight AI artifacts visible only upon close inspection, largely stable, well-matched lighting and colour, and believable physical interactions.
5 - Flawless quality: perfectly stable and temporally coherent, completely seamless integration, physically correct throughout, and natural motion. The edit is indistinguishable from the original footage.

# Dimension 5 Overall Performance

Provide a holistic assessment: considering all the above dimensions together, how well does the edited video meet the user's likely expectation? This score reflects overall subjective satisfaction - whether the result would be accepted by a real user as a successful reference-guided edit.

1 - Completely fails to meet user expectations. The edit is unusable.
2 - The edit attempt is recognisable but the result is clearly unacceptable due to combined failures across dimensions.
3 - A partially successful edit: the intent is understood and partially achieved, but noticeable issues in one or more dimensions reduce the result's usability.
4 - A good edit that would satisfy most users. Minor imperfections exist but do not significantly detract from the overall result.
5 - An excellent edit that fully meets or exceeds user expectations. The result is professional-grade, faithfully reflects the reference, and is virtually indistinguishable from a real or manually crafted video.

# Scoring Constraints

- If the instruction is not executed at all (non-execution), Instruction Compliance should be scored 1, and Video Consistency, Reference Image Consistency, and Generation Quality should be skipped (marked as "N/A"). Overall Performance should be 1.

# Response Format

Please output a valid JSON object exactly matching the following structure, without any extra markdown formatting or conversational text:
{{
  "Brief reasoning": "<A concise explanation covering all five dimensions, no more than 60 words.>",
  "Instruction Compliance": "<1-5>",
  "Video Consistency": "<1-5 or \"N/A\">",
  "Reference Image Consistency": "<1-5 or \"N/A\">",
  "Generation Quality": "<1-5 or \"N/A\">",
  "Overall Performance": "<1-5>"
}}

editing instruction is: {edit_prompt}.
Below are the original video, reference image, and edited video:
\end{lstlisting}

\section{Experimental Results}

\subsection{More Qualitative Comparison with SoTA Methods}
\label{app:more_qualitative}
\begin{figure}[htbp]
    \centering
    \includegraphics[width=1.0\linewidth]{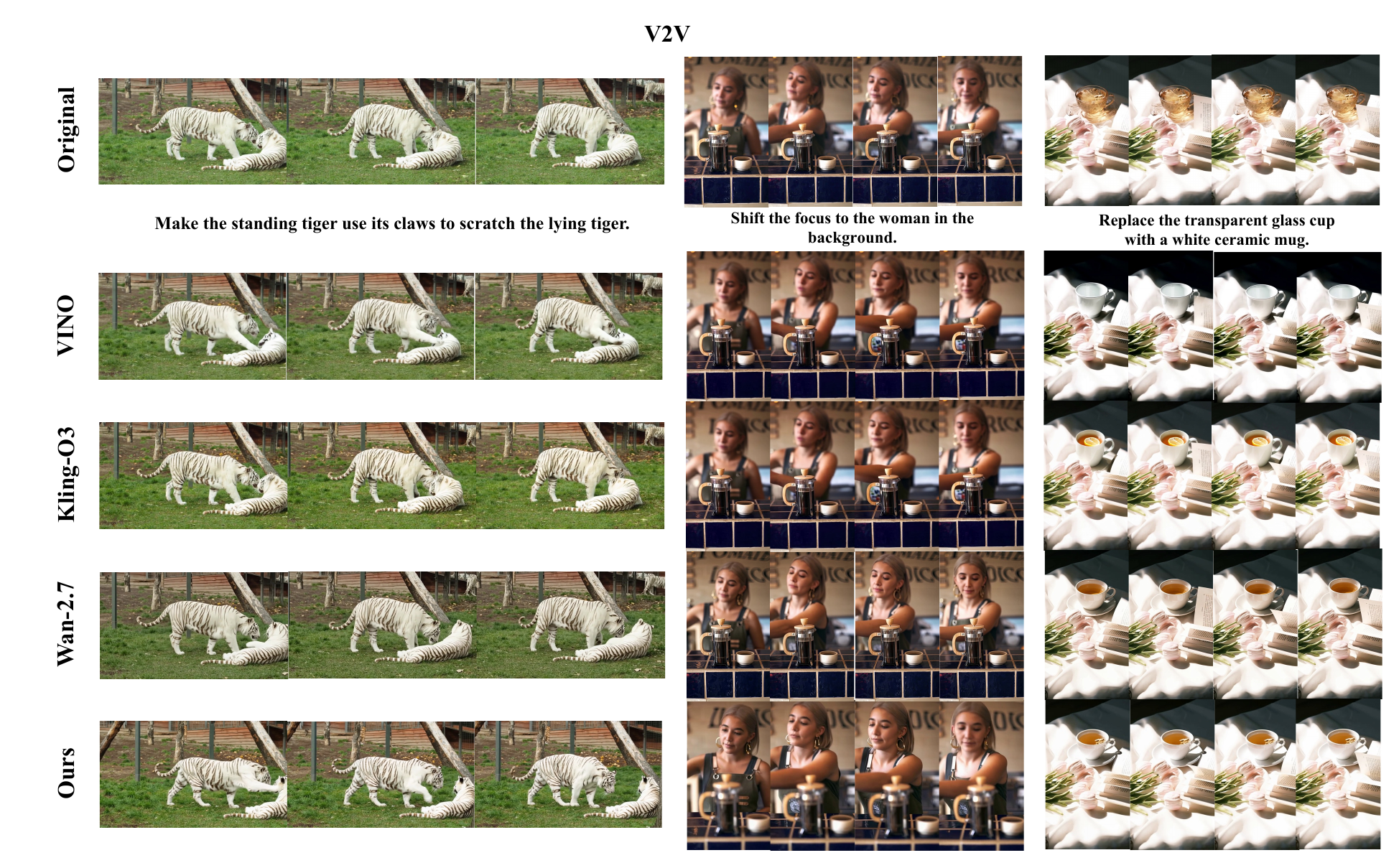}
    \caption{Qualitative Comparison with SoTAs on V2V task.}
    \label{fig:appendix_qualitative_v2v}
\end{figure}
\begin{figure}[t]
    \centering
    \includegraphics[width=1.0\linewidth]{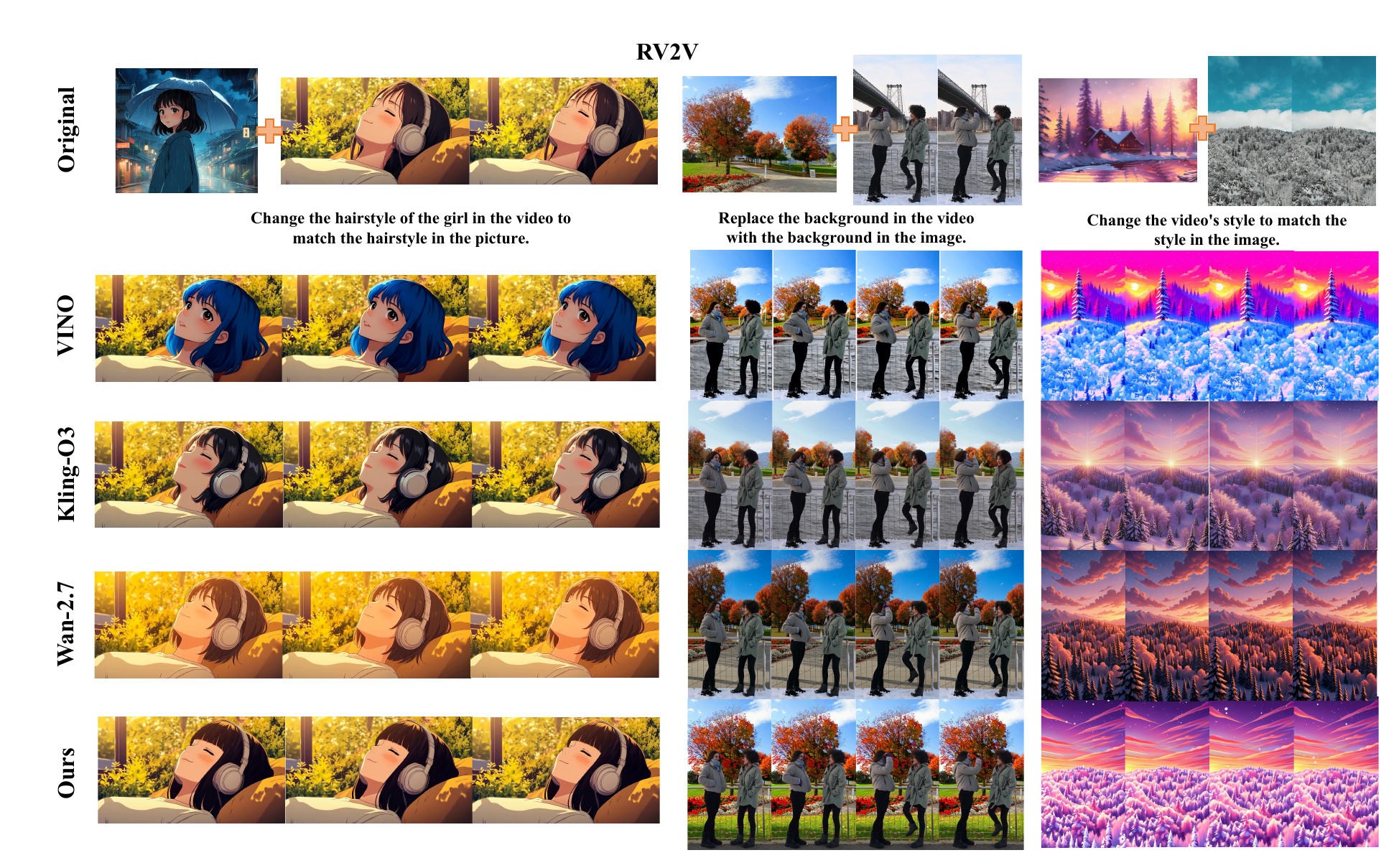}
    \caption{Qualitative Comparison with SoTAs on RV2V task.}
    \label{fig:appendix_qualitative_vr2v}
\end{figure}

Figure~\ref{fig:appendix_qualitative_v2v} shows additional results on V2V tasks. In the case on the left, neither Wan2.7 nor Kling O3 follows the instruction to modify the white tiger’s motion. VINO changes the motion, but the standing tiger keeps its paw on the lying tiger throughout the video, making the action look unnatural. In contrast, Bernini correctly produces the “scratching” motion. For the middle case, which involves shifting the focus, only Bernini moves the focus to the girl in the background while blurring the foreground. In the case on the right, which involves changing the material, all models successfully modify the material, but only Bernini preserves the tea and tea leaves inside the teacup.

Figure~\ref{fig:appendix_qualitative_vr2v} presents additional qualitative comparisons on the VR2V task. In the first case, both VINO and Kling alter the girl’s facial features, Wan2.7 changes the overall lighting of the video, while Bernini correctly modifies only the girl’s hairstyle. In the middle case, which involves changing the background, only Bernini removes the snow on the ground beneath the person’s feet from the original video. In the case on the right, only Bernini accurately applies the style of the reference image while preserving the original shape of the tree.

\subsection{Qualitative Comparison of Video Editing with Reasoning}
\label{app:vis_cot}
\begin{figure}[t]
\centering
\includegraphics[width=1\linewidth]{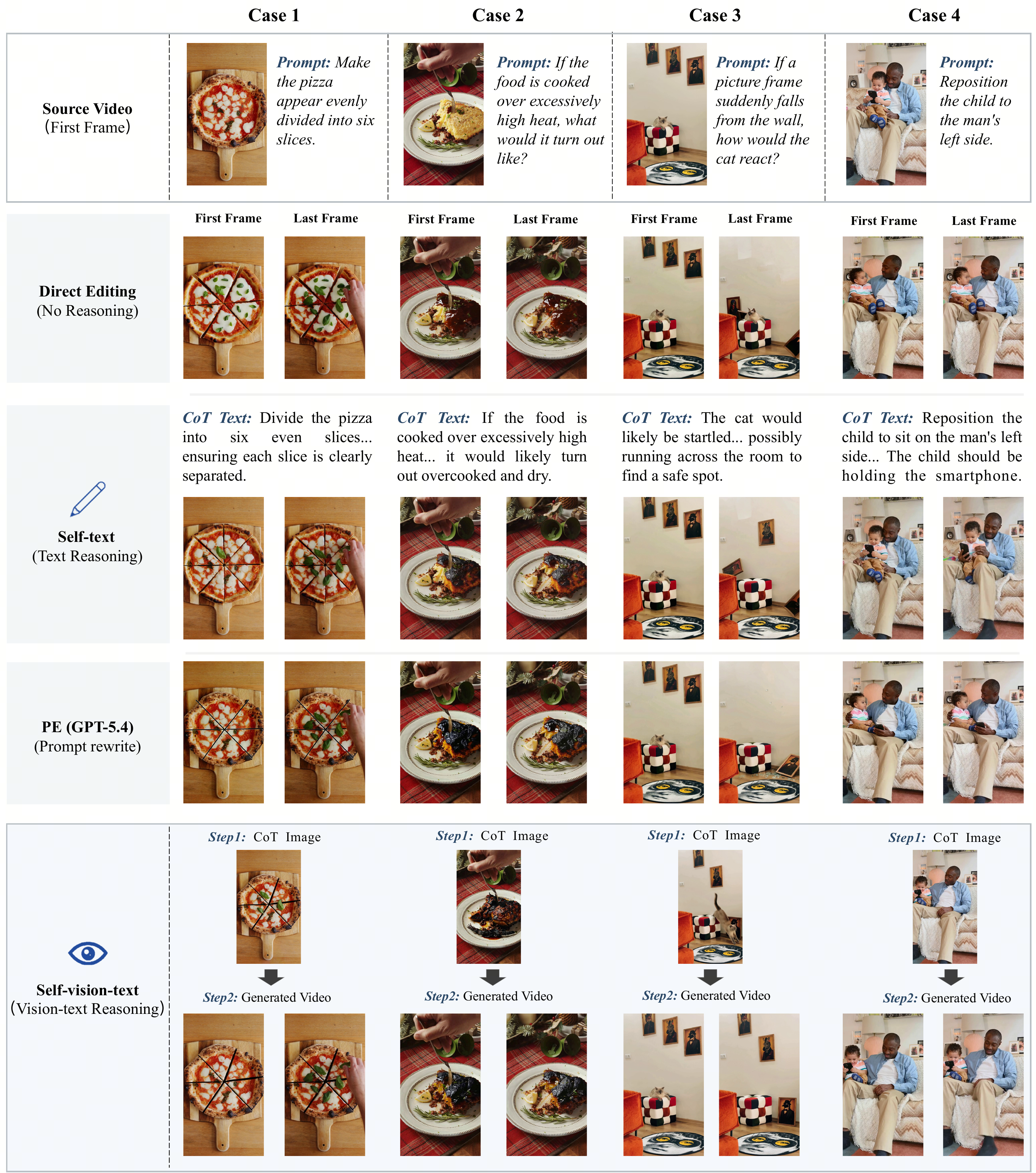}
\caption{Qualitative comparisons of different inference modes. For each video, we display the first and last frames. The methods are arranged from top to bottom, employing increasingly sophisticated reasoning patterns that lead to progressively improved editing quality. Our self-supervised vision-text reasoning further introduces a visual intermediate (the CoT image) to ground the process in the visual domain, yielding superior spatial fidelity and temporal consistency.}

\label{fig:cot_exp_vis}
\end{figure}

Figure~\ref{fig:cot_exp_vis} presents qualitative comparisons across different reasoning variants. The visualization reveals that while the baseline struggles with complex layout adjustments and imaginary state changes, incorporating self-generated textual reasoning yields notable improvements in instruction following.
Employing an LLM rewriter to refine input prompts further enhances performance, producing results with better structure and precision.
The intermediate CoT images provide essential visual grounding, ensuring superior alignment for challenging scenarios such as out-of-distribution tasks.

\subsection{More Generalization Results}
\begin{figure*}[t]
    \centering
    \includegraphics[width=1\linewidth]{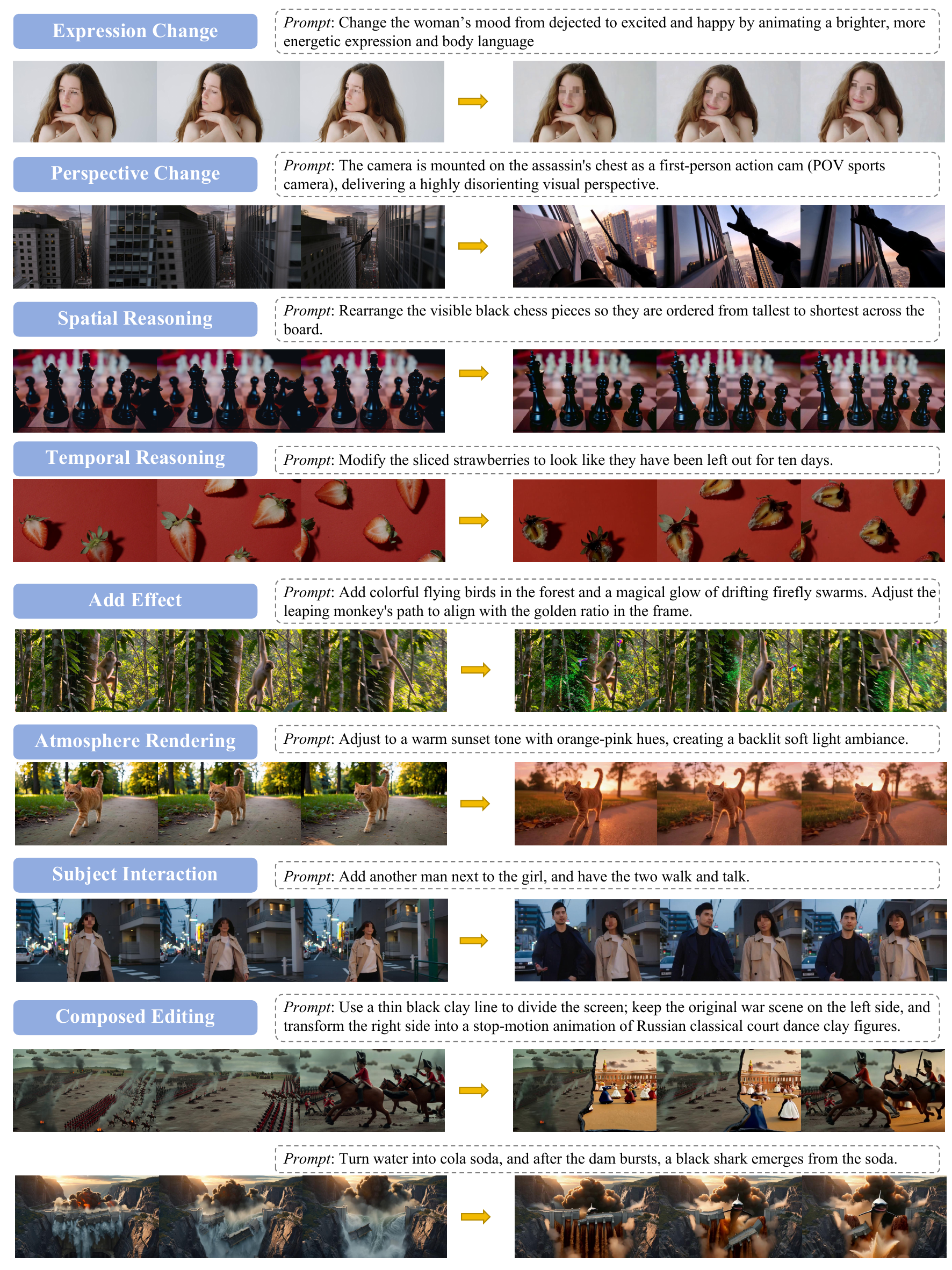}
    \caption{More generalization to diverse video editing instructions.}
    \label{fig:generalize_app}
\end{figure*}
Figure~\ref{fig:generalize_app} presents additional generalization examples on diverse video editing instructions beyond those covered during training, including expression change, perspective change, spatial reasoning, temporal reasoning, effect addition, atmosphere rendering, subject interaction, and composed editing. Although these editing types are absent from the V2V training data, Bernini is still able to execute them effectively. These results indicate that Bernini develops a transferable instruction-following capability that extends beyond the specific editing patterns observed during training.

\clearpage
\section{The EditingArena Leaderboard}
\label{app:editing-arena}

Public video-editing benchmarks are dominated by a handful of fixed
source clips and template instructions, which (i) saturate quickly,
(ii) reward methods that overfit to a known test distribution, and
(iii) collapse the many distinct kinds of editing into a single
scalar. Because Bernini targets instruction-following editing
that requires semantic planning rather than appearance matching, we
built \textbf{EditingArena}, a self-hosted, human-preference arena in
the spirit of LMSYS Chatbot Arena~\citep{chiang2024chatbot}, to
evaluate editing models under open-vocabulary, human-authored
instructions and blind pairwise judgment. This appendix
describes the collection protocol, the rating model, and the
diversity of the instruction set.

\subsection{Collection Protocol}
\label{app:arena-protocol}

Each arena episode is a three-stage pipeline. The key design choice is
that both the source video and the editing instruction are
freely authored by a human annotator, so the evaluation distribution is
never fixed in advance and cannot be memorized by any candidate.

\begin{enumerate}
  \item \textbf{Source generation (T2V).} The annotator writes a free-form
  text-to-video prompt, and a fixed text-to-video backend renders a
  source clip. Using a generation model rather than a static
  clip pool means the space of source content is effectively unbounded
  and identical across all methods being compared.

  \item \textbf{Blind editing (V2V).} The annotator writes an editing
  instruction. The system samples two distinct candidate methods
  from the enabled pool by weighted sampling without replacement and
  runs them concurrently on the same source clip. The two outputs
  are presented side by side as anonymous slots~\textsf{A}/\textsf{B},
  with the slot assignment randomized per round so that method identity
  stays uncorrelated with screen position.

  \item \textbf{Preference vote.} The annotator casts a single
  four-way vote: \textsf{A better}, \textsf{B better},
  \textsf{both good} (tie-good), or \textsf{both bad} (tie-bad).
  Method names are revealed only after the vote is recorded.
\end{enumerate}

\subsection{Rating Aggregation}
\label{app:arena-bt}

We aggregate the pairwise votes with a Bradley-Terry (BT) maximum
likelihood model~\citep{bradley1952rank}, the same family used by
LMSYS Chatbot Arena. Each method $i$ is assigned a latent strength
$s_i>0$ such that
\begin{equation}
  P(i \text{ beats } j) \;=\; \frac{s_i}{s_i + s_j}.
\end{equation}
A tie (either tie-good or tie-bad) is counted as half a
win plus half a loss for each side. We fit $\{s_i\}$ by the Zermelo
fixed-point iteration on the aggregated pair counts and normalize the
strengths to unit geometric mean, then report an Elo-anchored score
\begin{equation}
  R_i \;=\; 400 \cdot \log_{10}(s_i) \;+\; 1000,
\end{equation}
so that an ``average'' method scores $1000$. We obtain $95\%$
confidence intervals by a multinomial bootstrap: for each method pair
we redraw its observed votes from the empirical
$(\text{left-win},\text{right-win},\text{tie})$ distribution, refit BT,
and take the $2.5/97.5$ percentiles over $1000$ resamples. Alongside
the BT score we also report the raw win/loss/tie record and the full
pairwise win-rate matrix, so that head-to-head results can be read off
directly without trusting the BT model.

\subsection{Instruction Diversity}
\label{app:arena-diversity}

Instructions are authored in natural language (Chinese) by the
annotators with no template, so they vary in length, specificity, and
compositionality, ranging from terse single-edit commands
(``remove the pony'') to multi-clause directives that
combine an object edit, a lighting change, and a camera move. To
characterize coverage we post-hoc cluster the collected instructions
into thirteen editing intents spanning four broad axes
(Table~\ref{tab:arena-taxonomy}): local object edits, global appearance
edits, camera/geometry edits, and edits that require world knowledge and
temporal reasoning rather than mere appearance transfer. The last axis is
the most relevant to Bernini's semantic-planning thesis, for example
``show how this cup of coffee looks after sitting for three hours.''

\begin{table}[t]
  \centering
  \small
  \caption{Taxonomy of editing intents observed in EditingArena. The
  set spans purely visual edits as well as edits requiring world
  knowledge / temporal reasoning, which probe instruction
  understanding rather than texture transfer.}
  \label{tab:arena-taxonomy}
  \begin{tabular}{llp{6.6cm}}
    \toprule
    Axis & Intent & Description \\
    \midrule
    \multirow{4}{*}{\shortstack[l]{Local\\object}}
      & Subject replacement & swap one entity / background for another \\
      & Subject addition    & insert a new entity or agent into the scene \\
      & Subject removal     & delete an entity and inpaint behind it \\
      & Subject attribute   & change color / material / garment of an entity \\
    \midrule
    \multirow{4}{*}{\shortstack[l]{Global\\appearance}}
      & Style transfer      & re-render in a target artistic style \\
      & Color grading       & global tone / saturation / exposure shift \\
      & Atmosphere \& light & add mist, glow, mood lighting \\
      & Weather change      & rain, snow, lightning, fog \\
    \midrule
    \multirow{2}{*}{\shortstack[l]{Camera /\\geometry}}
      & Camera motion       & static$\leftrightarrow$dolly / pan / follow \\
      & Viewpoint change    & change shot angle (e.g.\ to bird's-eye / orbit) \\
    \midrule
    \multirow{2}{*}{\shortstack[l]{Knowledge /\\reasoning}}
      & Expression \& emotion & change a subject's facial expression / affect \\
      & Temporal reasoning    & roll the scene forward/backward in time, show a process \\
    \midrule
    -- & Other / compound    & combinations and edits outside the above \\
    \bottomrule
  \end{tabular}
\end{table}

\subsection{Representative Instruction Cases}
\label{app:arena-cases}

Table~\ref{tab:arena-cases} samples instructions across this range. The
upper block lists single, self-contained edits, one per intent. The
lower blocks move to instructions that combine several operations in a
single sentence, ask for imaginative transformations or narrative
re-staging, or call for world knowledge and temporal reasoning about how
a scene evolves. Many of the latter target a state that appearance
transfer alone cannot reach, such as how a cup of coffee settles after
three hours, so a method must first parse the request and plan a
multi-part edit before rendering it. Instructions were authored in
Chinese and are translated here for readability.

\begin{table}[t]
  \centering
  \small
  \caption{Representative editing instructions from EditingArena
  (translated from Chinese). The upper block gives one single-intent
  example per category, and the lower blocks show instructions that
  combine multiple operations or call for reasoning about how a scene
  changes.}
  \label{tab:arena-cases}
  \begin{tabular}{p{1.9cm}p{10.8cm}}
    \toprule
    Type & Instruction \\
    \midrule
    \multirow{8}{1.9cm}{\raggedright Single intent}
      & ``Change the background to Mars.'' \\
      & ``Add three flying seagulls in the background sky.'' \\
      & ``Remove the monk and keep everything else unchanged.'' \\
      & ``Change the girl's white dress to bright blue.'' \\
      & ``Turn it into a Chinese ink-wash landscape with negative-space composition.'' \\
      & ``Change the static shot into a slow forward dolly-in.'' \\
      & ``Change the side view into a top-down bird's-eye view.'' \\
      & ``Change the old man's expression from stern to smiling.'' \\
    \midrule
    \multirow{3}{1.9cm}{\raggedright Composite}
      & ``Remove the magician but keep the water rapidly freezing, while a flaming meteor crashes into the ice and the time shifts to deep night.'' \\
      & ``Turn the sunny scene into a downpour, with puddle reflections on the ground, people and clothes looking soaked, and the background trees veiled in rain mist.'' \\
      & ``Change the weather outside the window to rain, the time to dusk, and the palette to a cool blue-purple, and lower the saturation.'' \\
    \midrule
    \multirow{3}{1.9cm}{\raggedright Imaginative / narrative}
      & ``The white van transforms into a robot mid-drive and fires at a helicopter, then the frame turns black-and-white with raised contrast.'' \\
      & ``Add a giant cartoon juicer to the village. The partying cartoon vegetables line up, jump into the juicer and turn into cups of juice, and the camera moves to a static wide shot.'' \\
      & ``Make the train powered by several white horses galloping along the track to pull it, rendered in 2D Disney Alice-in-Wonderland style.'' \\
    \midrule
    \multirow{4}{1.9cm}{\raggedright Knowledge / reasoning}
      & ``Show this coffee after sitting three hours, with a film on the surface, dried stains on the rim, no steam left, and the light shifting from daylight to warm evening lamplight.'' \\
      & ``Replace the cake with how it looks after three days, the surface slightly dried out and the color a little darker.'' \\
      & ``Show this firefly's metamorphosis from larva to adult.'' \\
      & ``What does it look like after the blizzard?'' \\
    \midrule
    \multirow{3}{1.9cm}{\raggedright Fine-grained}
      & ``Change the phone from black to white, raise the screen brightness by 5\%, and keep the text legible.'' \\
      & ``Change the clear water to orange juice, add more bubbles, and keep the glass transparent.'' \\
      & ``Delete the snowman's legs and feet but keep the footprints, then convert the frame to a classical realist oil-painting style.'' \\
    \bottomrule
  \end{tabular}
\end{table}

\clearpage
\section{Segment-Index Interpolation for More Reference Inputs}
\label{app:src_id_interp}

SA-3D RoPE is trained with a bounded number of visual segments: the target
segment takes index $i=0$ and the conditioning segments take integer indices up
to $i_{\max}$ ($i_{\max}=5$ in our setting). At inference, however, a request may
carry more reference inputs than seen during training. Naively assigning the
extra segments indices $i > i_{\max}$ extrapolates the segment phase
$\mathbf{r}^{\mathrm{seg}}_i$ outside the trained manifold, a regime the renderer
was never optimized for. In practice this degrades reference consistency and
yields pronounced artifacts: the rendered subjects become faded and
semi-transparent, the background bleeds through the foreground, and the model can
no longer reliably bind each conditioning segment to its target content (severe
ghosting and identity loss).

Since the rotary phase is a continuous function of the segment index, fractional
indices remain well-defined. When $n > i_{\max}$ conditioning segments are given,
we therefore remap their indices evenly into the trained range,
\begin{equation}
i_k = 1 + \frac{(k-1)\,(i_{\max}-1)}{n-1}, \qquad k = 1, \dots, n,
\label{eq:src_id_interp}
\end{equation}
while the target segment keeps $i=0$; for $n \le i_{\max}$ the integer indices are
kept unchanged. This mirrors positional interpolation for context-window extension in LLMs~\cite{chen2023extending,peng2023yarn,ding2024longrope}: instead of asking the model to generalize to unseen rotary phases, we keep all phases within the trained distribution.

\begin{figure*}[t]
  \centering
  \includegraphics[width=\linewidth]{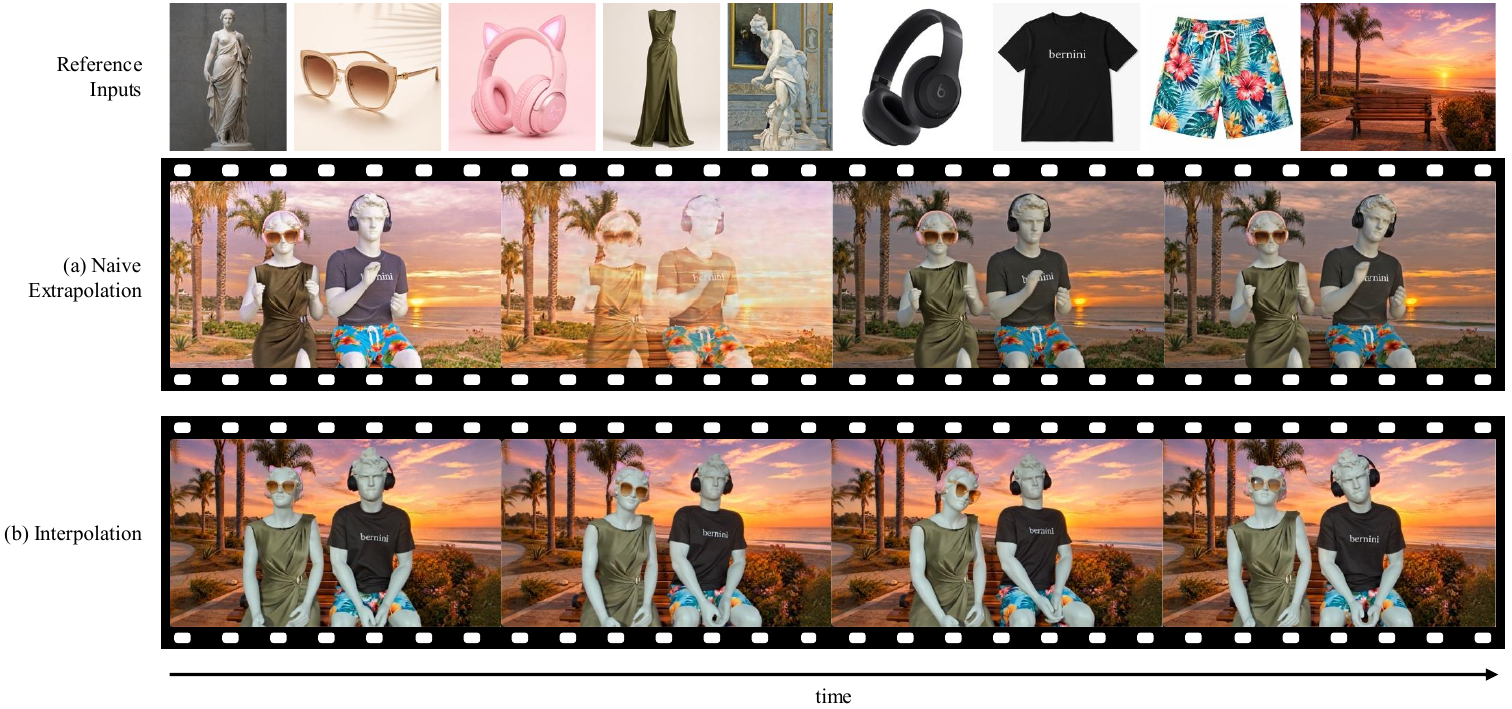}
  \caption{Segment-index interpolation vs.\ naive extrapolation when the number
  of reference inputs exceeds the trained maximum ($n > i_{\max}$). \textbf{(a)}
  With naively extrapolated indices ($i > i_{\max}$), the renderer produces
  faded, semi-transparent subjects with ghosting and loss of reference identity.
  \textbf{(b)} Remapping the segment indices into the trained range
  (Eq.~\ref{eq:src_id_interp}) keeps each reference bound to its target region
  and preserves identity throughout the clip.}
  \label{fig:src_id_interp}
\end{figure*}

Fig.~\ref{fig:src_id_interp} compares interpolation against naive extrapolation
with $n=9$ reference inputs. 
With extrapolated indices, the rendered subjects become faded and
semi-transparent, with strong ghosting through which the beach background is
visible, and the references are no longer correctly bound to their target
regions. With interpolation, the renderer keeps each reference correctly bound to
its target region and preserves subject identity and opacity throughout the clip.
Note that interpolation compresses the phase gap between adjacent segments (from
$1$ to $(i_{\max}-1)/(n-1)$), so segment separability may degrade for very large
$n$; in practice we verified the behavior up to $n=10$.

\end{document}